\documentclass{article}
\usepackage{iclr2027_conference,times}

\usepackage{amsmath,amsfonts,bm}

\def\eqref#1{equation~\ref{#1}}
\def\1{\bm{1}}

\DeclareMathAlphabet{\mathsfit}{\encodingdefault}{\sfdefault}{m}{sl}
\SetMathAlphabet{\mathsfit}{bold}{\encodingdefault}{\sfdefault}{bx}{n}

\usepackage{hyperref}

\usepackage{amsmath}
\usepackage{amssymb}
\usepackage{mathtools}
\usepackage{amsthm}

\usepackage[table]{xcolor} 

\usepackage{tcolorbox}
\tcbuselibrary{skins}

\usepackage{multirow}
\usepackage{array}
\usepackage{longtable}
\usepackage{booktabs}

\usepackage[linesnumbered,ruled,vlined,algo2e]{algorithm2e}
\usepackage{enumitem}
\usepackage{xurl}

\usepackage{arydshln}
\usepackage{wrapfig}

\definecolor{BlueGreen}{RGB}{0,128,128}
\definecolor{RedOrange}{RGB}{255,69,0}
\definecolor{softpurple}{HTML}{B49ABF}
\definecolor{softorange}{HTML}{FFA816}
\definecolor{softblue}{RGB}{121,151,161}
\definecolor{mygray}{gray}{0.4}

\usepackage{environ}

\title{MASRubric: Auditing Information Flow in Multi-Agent Systems with Failure-Distilled Pitfall Rubrics}

\author{Yutong Wang$^1$\thanks{~Equal Contribution.}~
        Siyuan Xiong$^{1*}$~
        Xuebo Liu$^1$\thanks{~Xuebo Liu is the corresponding author.}~
        \textbf{Wenkang Zhou}$^1$\thanks{~This work was done when Wenkang Zhou was interning at Harbin Institute of Technology, Shenzhen.}~
        \textbf{Liang Ding}$^2$~\\
        \textbf{Miao Zhang}$^1$~
        \textbf{Min Zhang}$^1$~ \\
        $^1$Harbin Institute of Technology, Shenzhen \;
        $^2$Alibaba Group \\
        \texttt{\{wangyutong,xiongsiyuan\}@stu.hit.edu.cn} \\
        \texttt{\{liuxuebo,zhangmiao,zhangmin2021\}@hit.edu.cn} \\
        \texttt{zhouwenkang22@mails.ucas.ac.cn} \;
        \texttt{liangding.liam@gmail.com} \\
}

\iclrfinalcopy
\begin{document}

\maketitle

\begin{abstract}
While multi-agent systems (MAS) excel at complex reasoning, they are vulnerable to errors that intermediate agents introduce and downstream agents build upon.
Auditing intermediate messages before they propagate requires an explicit standard, yet evaluation rubrics are typically authored by domain experts or written against a reference answer, neither of which is available for an unseen message at test time.
We present \textbf{MASRubric}, a MAS information flow auditing framework with failure-distilled pitfall rubrics.
Offline, trajectories on which the MAS has failed are automatically distilled into a reusable bank of pitfall criteria, each describing a recurrent error by its underlying misconception, the reasoning situations in which it arises, and the check that would expose it.
Online, the criteria applicable to each intermediate message are retrieved from this off-the-shelf bank and checked one by one, and the resulting satisfaction rate decides whether the message is broadcast, returned to its author with diagnostic feedback for revision, or withheld.
Empirical results demonstrate that MASRubric enhances MAS performance on both fixed and dynamic frameworks, achieving average accuracy gains of up to 2.83 points on math reasoning benchmarks and 1.74 points on code generation benchmarks.
Further analysis shows that the retrieved criteria vary systematically with task types, and that the audit effort tracks task difficulty.
Moreover, the bank transfers without re-mining to a system with a stronger backbone, which makes more adaptive and more efficient use of it.
Our code and dataset are released at \url{https://github.com/TonySY2/MASRubric}.
\end{abstract}

\section{Introduction}
\begin{wrapfigure}{r}{0.5\textwidth}
    \vspace{-15pt}
    \centering
    \includegraphics[width=\linewidth]{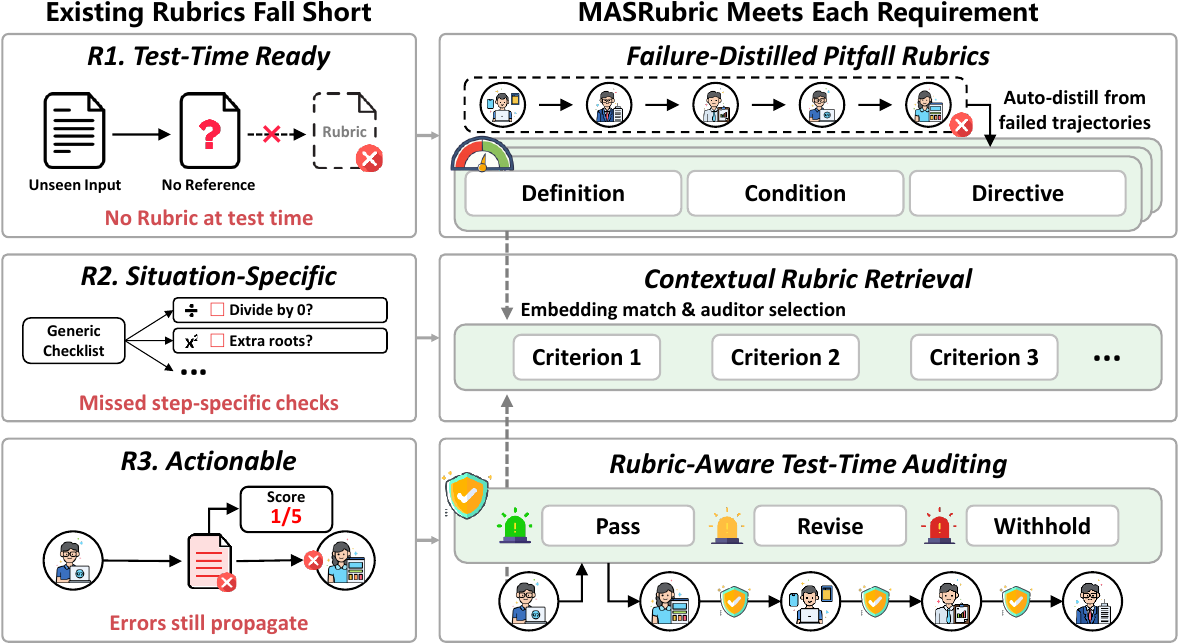}
    % \vspace{-18pt}
    \caption{Existing rubrics (left) versus MASRubric (right) on each requirement.}
    \label{fig:intro}
    \vspace{-10pt}
\end{wrapfigure}
Large language model (LLM)-based multi-agent systems (MAS) achieve remarkable performance on complex tasks such as software development \citep{hong2024metagpt}, ultra-long context handling \citep{li-etal-2024-graphreader}, and scientific discovery \citep{https://doi.org/10.1002/adma.202413523}, by harnessing collective intelligence \citep{zhuge2024gptswarm, wang2026maspo} and orchestrating cooperative teams \citep{zhang2025gdesigner, zhang2025evoflowevolvingdiverseagentic}.
Collaboration, however, also opens a channel for failure.
An intermediate message enters the context of every downstream agent without being checked, so a locally plausible error is taken as settled evidence, elaborated upon, and eventually turns into a task-level failure \citep{zhang2025which}.
Thus, the benefit of collaboration is not guaranteed: MAS have been reported to offer only marginal gains over, or even to trail, a well-prompted single agent \citep{wang-etal-2024-rethinking-bounds, pan2025why}.
What needs to be checked in a MAS is thus the intermediate message, not only the final answer.

Auditing intermediate messages requires a standard to audit them against, and a useful standard must meet three requirements (Figure \ref{fig:intro}).
First, it should be \emph{available at test time} (R1): it must apply to an unseen input, without a reference answer and without anyone writing it on the spot.
Second, it should be \emph{situation-specific} (R2): what is worth checking depends on the reasoning an agent performs at that step, and a uniform checklist either misses step-specific pitfalls or burdens every message with irrelevant checks.
Third, its verdict should be \emph{actionable} (R3): it must change what is passed downstream, rather than merely scoring a message after the fact.
Existing mechanisms each miss at least one of these.
Self-refine appeals to no standard beyond the reasoner's own judgment \citep{madaan2023selfrefine, shinn2023reflexion}.
Standardized operating procedures and guardrail policies provide explicit standards, but these are hand-written for each domain and applied uniformly to every message \citep{hong2024metagpt, xiang2025guardagent}.
Rubrics are the most natural form for explicit criteria, yet they are typically written against a reference answer to describe what a correct answer should contain, so they cannot be instantiated for an unseen input, and their verdicts are consumed as scores or rewards \citep{arora2025healthbenchevaluatinglargelanguage, gunjal2026rubrics}.

To meet these requirements, we propose \textbf{MASRubric} (Figure \ref{fig:framework}).
Offline, trajectories on which the MAS has failed are distilled into a reusable bank of \emph{pitfall criteria}, each stating what a given kind of reasoning is likely to get wrong, the situation in which that risk arises, and the check that would expose it.
Because they describe pitfalls rather than the content of a correct answer, the criteria transfer to unseen queries; reference answers are used only offline, to tell failed runs from successful ones, so the bank applies at test time with no reference answer and no human-written criterion (R1).
Online, the criteria in the off-the-shelf bank, whose applicability conditions match the current reasoning situation are retrieved for each intermediate message, forming a context-specific rubric (R2).
The auditing agent then judges the message criterion by criterion, and the verdict decides whether it is passed to broadcast, returned for revision, or withheld (R3).
Across a fixed-topology and a dynamically routed MAS on nine mathematical and four code generation benchmarks, MASRubric improves the average accuracy of the underlying MAS in every setting tested, by 1.59 and 2.83 points on mathematics for the two frameworks and by 1.74 points on code, with the largest gains on olympiad-level problems.
A bank mined from a Qwen3-8B system also transfers unchanged to a Qwen3-14B system (52.56\% $\rightarrow$ 56.29\%), whose stronger backbone models allocate criteria more sharply by difficulty while needing fewer revision rounds.
Our contributions are as follows:
\begin{itemize}
    \item We distill a MAS's sampled failure trajectories offline into an \textit{off-the-shelf} bank of reusable pitfall criteria with explicit applicability conditions, from which pitfall rubrics can be assembled at test time without any reference answer or human-written criterion.
    \item We design a rubric-aware auditing framework for MAS that retrieves a \textit{context-specific rubric} for each agent's message by matching applicability conditions to its reasoning situation, and turns criterion-level verdicts into revision or withholding before errors cascade.
    \item We find that the most-retrieved criteria vary with task type and difficulty, so the bank supplies \textit{distinct pitfalls} for distinct situations rather than a generic checklist. Audit rounds and withholding rates also track task difficulty, suggesting a reference-free difficulty signal.
\end{itemize}

\section{Related Work}

\textbf{Multi-Agent Systems and Error Propagation.}
LLM-based MAS, applied to software development \citep{hong2024metagpt, qian-etal-2024-chatdev}, scientific discovery \citep{https://doi.org/10.1002/adma.202413523, ghareeb2026multi}, and long-context understanding \citep{li-etal-2024-graphreader, zhao-etal-2024-longagent}, increasingly organize themselves automatically, searching communication topologies as optimizable graphs \citep{zhuge2024gptswarm, zhang2025gdesigner, zhang2025aflow} or assembling specialized teams per task \citep{tian-etal-2025-agentinit, dang2025multiagent, wang2026orchestratingintelligenceconfidenceawarerouting}.
Scaling collaboration, however, also scales exposure to a single faulty participant \citep{liu2026propuqmaspropagationawareuncertaintyquantification}.
Failure-attribution studies trace such breakdowns to particular agents and steps \citep{zhang2025which, pan2025why, zhang2026agentracer, qiao2026verifymashypothesisverificationfailure}, and intervention-based debugging shows that targeted edits to intermediate messages or plans can turn failed runs into successes \citep{ma2026dover}.

\textbf{Test-Time Defenses against Cascading Errors.}
Since retraining agents for every deployment is impractical, many defenses act during inference.
Self-refine asks an agent to revise its own output before passing it downstream \citep{madaan2023selfrefine, shinn2023reflexion, paul-etal-2024-refiner}, without any external standard.
External monitors instead screen messages against explicit verification policies \citep{xiang2025guardagent, 3780338.3780653, luo-etal-2025-agrail} or flag anomalies in the communication graph \citep{wang-etal-2025-g, zhou2025guardian}.
A third line builds checkability into the design itself, either imposing standardized operating procedures so that agents exchange structured, verifiable artifacts instead of free-form messages \citep{hong2024metagpt, qian-etal-2024-chatdev}, or restricting who may transmit to whom \citep{zhang2025cut, wang-etal-2025-agentdropout}.
Unlike these standards, which are hand-specified or applied uniformly to every message, MASRubric mines its standard from execution evidence and selects it per message.

\textbf{Rubric Construction and Use.}
Rubrics make evaluation explicit by decomposing a verdict into named criteria, and differ mainly in where those criteria come from: expert authoring \citep{arora2025healthbenchevaluatinglargelanguage}, per-query generation from a reference answer aggregated into a reward \citep{gunjal2026rubrics}, distillation from preference pairs compressed by an information-theoretic coding rate \citep{xie2026autorubriclearningimplicitweights}, or retrieval of similar queries' rubrics to generate an instance-specific one at inference time \citep{10.1145/3805712.3809940}.
Closer to our setting, rubrics score transitions for credit assignment during training \citep{zhang2026trcatransitionwiserubriccredit} and verify candidate solutions at test time, where an agent generates them per instance by exploring the repository \citep{raghavendra-etal-2026-agentic}.
Ours are distilled instead from the system's own failures without human authoring, and state what the reasoning at hand is likely to get wrong rather than what a correct answer should contain, which lets them apply to messages that have no reference of their own.
Their verdicts, moreover, decide what content reaches downstream agents rather than yielding a score or a reward.

\section{Methodology}

\subsection{Formulation}\label{sec:formulation}

\textbf{Multi-Agent Workflow.}
A MAS execution activates $N$ agents in the execution order $\mathcal{S}=(A_{1},A_{2},...,A_{N})$ induced by the system's topology. Each agent is a tuple $A_{i}=(\Phi_{i},\mathcal{R}_{i},\mathcal{K}_{i})$: a backbone model $\Phi_{i}(\cdot)$ serving as the reasoning engine, a static role specification $\mathcal{R}_{i}$, and a dynamic knowledge base $\mathcal{K}_{i}$ holding the messages it can observe (initially $\mathcal{K}_{i}=\emptyset$). Given an input $x_{i}$, agent $A_{i}$ produces an output
$o_{i}=\Phi_{i}(x_{i},\mathcal{R}_{i},\mathcal{K}_{i})$,
which is then transmitted to every designated successor agent to update its knowledge base via:
\begin{equation}\label{eq:broadcast}
\mathcal{K}_{j}\leftarrow\mathcal{K}_{j}\cup\{(\mathcal{R}_{i},o_{i})\}.
\end{equation}
A run thus yields a complete inference trajectory $\mathcal{T}=(\mathcal{Q},A_{1:N},o_{1:N},\mathcal{Y})$ over the task query $\mathcal{Q}$, whose answer $\mathcal{Y}=o_{N}$ is produced by the terminal aggregator agent.
Equation \ref{eq:broadcast} places no condition on either whether a message is transmitted or what it contains: every intermediate message is passed on, exactly as written.
The content that accumulates in $\mathcal{K}_j$ is therefore whatever the upstream agents produce, which is precisely the channel along which a local mistake escalates into a task-level failure.

\begin{figure*}
    \centering
    \includegraphics[width=0.88\linewidth]{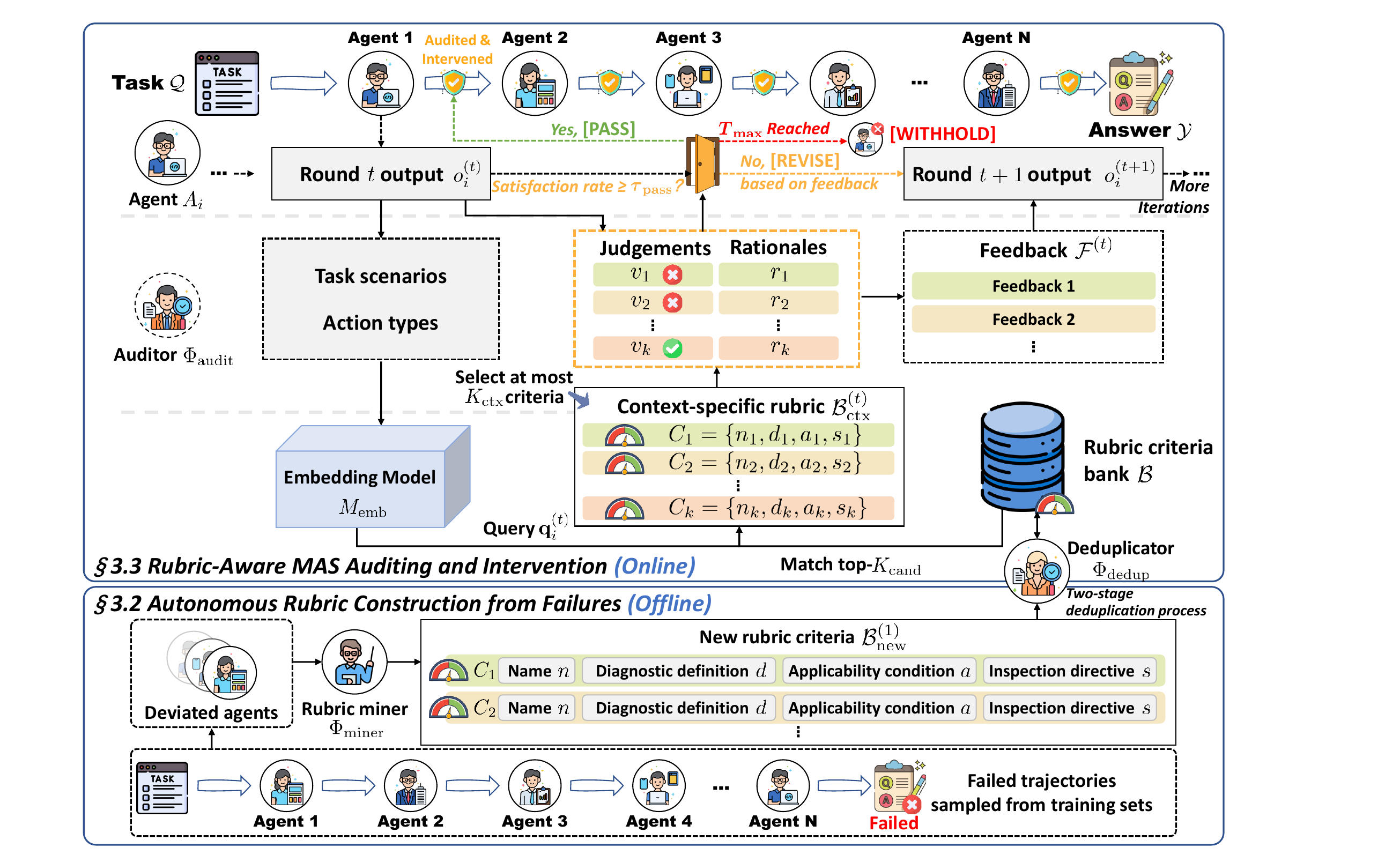}
    \caption{Overview of \textbf{MASRubric}. The lower block shows the offline stage, where failure trajectories of a MAS are distilled by a rubric miner into criteria and compacted by dual-stage redundancy elimination into a reusable criterion bank. The upper block shows the online stage, where a context-specific rubric is retrieved for each intermediate message, the auditing agent judges the message criterion by criterion, and the resulting satisfaction rate decides whether the message passes and is broadcast, is revised by its author with diagnostic feedback, or is withheld.}
    \label{fig:framework}
\end{figure*}

\textbf{Pitfall Rubrics.}
Containing this escalation requires auditing each message before it updates $\mathcal{K}_j$.
Self-refine, the common practice \citep{madaan2023selfrefine, shinn2023reflexion}, offers no standard beyond the reasoner's own judgment and is thus prone to confirmation bias.
We therefore audit against explicit criteria in the checklist style common in fine-grained LLM evaluation \citep{arora2025healthbenchevaluatinglargelanguage, gunjal2026rubrics}: each criterion is judged satisfied or violated, and verdicts are aggregated by counting.
Unlike rubrics that describe what a good answer should contain, our criteria are stated \emph{negatively}, each naming a recurrent reasoning pitfall, the situation in which it occurs, and the check that would expose it.
Formally, a rubric criterion is a structured object
\begin{equation}\label{eq:criterion}
    C = (n, d, a, s),
\end{equation}
whose fields separate \emph{what the failure mode is} from \emph{when it is relevant} and \emph{what has to be verified}.
The identifier $n$ names the failure mode; the diagnostic definition $d$ characterizes its root cause, i.e., the misconception that produces it; the applicability condition $a$ specifies the class of reasoning situations in which the criterion is relevant; and the inspection directive $s$ states the specific check to run on a message once the criterion applies.
The collection of all criteria forms a \textbf{rubric criterion bank} $\mathcal{B}=\{C_1,\dots,C_{|\mathcal{B}|}\}$, which is constructed offline and held fixed during inference.

\subsection{Autonomous Rubric Construction from Failures}\label{sec:bank_construction}

Just as mature organizations rely on institutional memory, which codifies lessons learned from past projects to prevent the recurrence of known pitfalls, our framework requires a structured repository of the ways in which a MAS actually goes wrong.
Hand-written criteria are costly to produce and capture only the failures their authors anticipate, while generic quality checklists are too coarse to pinpoint the errors that matter.
We therefore derive them from evidence: the trajectories on which the system has already failed (lower block of Figure \ref{fig:framework}).

\textbf{Offline Rubric Mining.}
Given a source dataset $\mathcal{D}_\text{src} = \{\mathcal{Q}, \mathcal{Y}^*\}$ of queries paired with ground-truth answers, we run the MAS on each query and collect the trajectories $\mathcal{T}$ whose answer $\mathcal{Y}$ differs from $\mathcal{Y}^*$ into a failure set $\mathcal{D}_\text{fail}$.
A rubric miner $\Phi_\text{miner}$ then scrutinizes individual agents within $\mathcal{D}_\text{fail}$.
Upon detecting a deviation in agent $A_i$'s output $o_i$ given its role $\mathcal{R}_i$ and the query $\mathcal{Q}$, $\Phi_\text{miner}$ synthesizes a set of criteria, each populated along the four fields of Equation \ref{eq:criterion}:
\begin{equation}\label{eq:mining}
    \mathcal{B}_\text{new}^{(i)} = \left\{C_1,\dots,C_{m_i}\right\} = \Phi_\text{miner}\left(\mathcal{T}, \mathcal{Y}^*, \mathcal{R}_i, o_i \right),
\end{equation}
where the number of mined criteria $m_i$ is decided by the miner itself and is capped at a small constant to discourage over-reporting; $m_i=0$ when the output is judged sound within the scope of its role, in which case $A_i$ contributes nothing to the bank even though the trajectory as a whole failed.
The candidates accumulated over all agents and all failed trajectories, $\bigcup_{\mathcal{T}\in\mathcal{D}_\text{fail}}\bigcup_{i}\mathcal{B}_\text{new}^{(i)}$, are then filtered into an off-the-shelf bank $\mathcal{B}$ by the procedure below.

\textbf{Redundancy Elimination.}
Failure trajectories repeat themselves, and a bank that mirrors this repetition would let near-duplicate criteria crowd out the limited number of slots available to the online rubric, narrowing the effective coverage of each audit.
Compactness of a reusable criterion set is thus a first-class design goal.
We therefore admit candidates one at a time, subjecting each newly generated $C_\text{new}$ to a two-stage deduplication process.
First, we encode the concatenation of its diagnostic definition and applicability condition into a semantic vector $\mathbf{v}_\text{new} = M_\text{emb}(d_\text{new} \oplus a_\text{new})$ to retrieve the $K_\text{dedup}$ most similar existing criteria $\mathcal{B}_\text{sim} \subset \mathcal{B}$ via cosine similarity.
Subsequently, a deduplication LLM $\Phi_\text{dedup}$ evaluates $C_\text{new}$ against $\mathcal{B}_\text{sim}$, appending it to $\mathcal{B}$ only if it represents a distinct error pattern.
Examples of the resulting criteria are provided in Appendix \ref{app:prompt}, and the full procedure is given in Algorithm \ref{alg:bank_construction}.

\subsection{Rubric-Aware MAS Auditing and Intervention}\label{sec:intervention}

Assembling a rubric is only useful if its verdict changes the MAS's behavior.
A natural locus of intervention is Equation \ref{eq:broadcast}, which governs \emph{what content} downstream agents get.
This section describes how the rubric is assembled for a message, how criterion-level judgments are produced and aggregated, and how the verdict is converted into modifications (upper block of Figure \ref{fig:framework}).

\textbf{Contextual Rubric Retrieval.}
Most criteria in a domain-wide bank are irrelevant to any given reasoning step, so the rubric in force must be assembled per message.
Unlike retrieval-augmented rubric generation, which fetches the rubrics of semantically similar queries in order to write a fresh one for the instance at hand \citep{10.1145/3805712.3809940}, our criteria already carry explicit applicability conditions: assembly therefore requires no generation step and reduces to matching the current reasoning situation against those conditions.
To allocate the most pertinent criteria to an agent $A_{i}$ producing output $o_{i}^{(t)}$ at iteration $t$, we employ a coarse-to-fine strategy.
First, the auditing agent extracts the task scenarios and the proposed action types from the reasoning context, which are encoded into a query vector $q_{i}^{(t)}$ by an embedding model $M_{\text{emb}}$.
We then perform semantic matching against the applicability condition embeddings in $\mathcal{B}$ to retrieve the top-$K_\text{cand}$ candidates, forming $\mathcal{B}_\text{cand}^{(t)}$.
Embedding the condition rather than the definition keeps the retrieval key aligned with the question being asked at this stage, namely which pitfalls this situation exposes.
Subsequently, the auditing agent $\Phi_\text{audit}$ analyzes the agent's input, role, and output to filter $\mathcal{B}_\text{cand}^{(t)}$, with each candidate presented by its identifier, diagnostic definition, and applicability condition, producing a context-specific rubric $\mathcal{B}_\text{ctx}^{(t)}=\Phi_\text{audit}(\mathcal{B}_\text{cand}^{(t)}\mid x_i,\mathcal{R}_i,o_i^{(t)})\subseteq\mathcal{B}_\text{cand}^{(t)}$ of at most $K_\text{ctx}$ criteria.

\textbf{Criterion-Level Judgment.}
Given $\mathcal{B}_\text{ctx}^{(t)}$, the auditing agent inspects the output $o_i^{(t)}$ once per criterion.
For each $C_k = (n_k, d_k, a_k, s_k) \in \mathcal{B}_\text{ctx}^{(t)}$, the audit is instantiated with its applicability condition and inspection directive alone, and $\Phi_\text{audit}$ returns a binary violation flag $v_k^{(t)} \in \{0, 1\}$, which indicates whether the output violates the criterion ($v_k^{(t)}=1$) or satisfies it ($v_k^{(t)}=0$), together with a diagnostic rationale $r_k^{(t)}$:
\begin{equation}
\left( v_k^{(t)}, r_k^{(t)} \right) = \Phi_\text{audit}\left( o_i^{(t)} \mid x_i, \mathcal{R}_i, (a_k, s_k) \right).
\end{equation}

\textbf{Aggregation and Tri-State Gating.}
To judge the message as a whole, we aggregate the per-criterion verdicts into the \textbf{rubric satisfaction rate} $p_i^{(t)}$, the proportion of the criteria that the output satisfies:
\begin{equation}\label{eq:passrate}
    p_i^{(t)} = 1 - \sum\nolimits_{k} v_k^{(t)} \big/ |\mathcal{B}_\text{ctx}^{(t)}|.
\end{equation}
We then compare $p_i^{(t)}$ against a threshold $\tau_\text{pass}\in[0,1]$ to drive a tri-state decision on the message:
\begin{itemize}[leftmargin=*,nosep]
    \item \textbf{Pass:} If the satisfaction rate meets the threshold ($p_i^{(t)} \ge \tau_\text{pass}$), the output is accepted and broadcast to downstream agents: $o_i = o_i^{(t)}$.
    \item \textbf{Revise:} If the satisfaction rate is insufficient ($p_i^{(t)} < \tau_\text{pass}$) and the iteration count $t < T_\text{max}$, we collect the diagnostic rationales of the violated criteria into a feedback set ${\mathcal{F}^{(t)}=\{r_k^{(t)}\mid v_k^{(t)}=1\}}$, from which the original reasoning agent regenerates its output $o_i^{(t+1)}=\Phi_i(x_i,\mathcal{R}_i,\mathcal{K}_i,\mathcal{F}^{(t)})$.
    \item \textbf{Withhold:} If the satisfaction rate remains sub-threshold once the budget is exhausted ($p_i^{(T_\text{max})} < \tau_\text{pass}$), the message is withheld ($o_i = \emptyset$) to prevent the spread of erroneous content.
\end{itemize}
The message transmitted to the successor set $\mathcal{N}(A_i)$ is thus the first output that reaches the threshold within $T_\text{max}$ revisions, or $\emptyset$ if none does.
The pseudo-code of the online procedure is provided in Algorithm \ref{alg:intervention} in Appendix \ref{app:codes}, detailed prompt specifications are listed in Appendix \ref{app:prompt}, and a case study tracing one message through its audit procedure is presented in Appendix \ref{app:case}.

\section{Experiment}

\subsection{Experimental Setup}

\textbf{MAS Framework.}
To comprehensively evaluate our method, we implement it across two distinct MAS infrastructures:
(1) \textbf{Fixed-MAS}: Following \citet{wang-etal-2025-agentdropout}, agents are organized within a Directed Acyclic Graph (DAG), executing workflows statically according to a topological sort.
(2) \textbf{Dynamic-MAS}: Utilizing AutoGen's \citep{wu2024autogen} SelectorGroupChat\footnote{https://microsoft.github.io/autogen/stable/user-guide/agentchat-user-guide/selector-group-chat.html} framework, where a selector iteratively routes tasks within a globally broadcasted context to reach a final decision.

\textbf{Baselines.} 
We evaluate our method against several paradigms of test-time MAS intervention: 
(1) \textbf{Self-Refine Style} \citep{madaan2023selfrefine}: Prompting the reasoner agent to independently review and revise its output at each reasoning round. 
(2) \textbf{PRM-guided Search}: Utilizing Qwen2.5-Math-PRM-7B \citep{zhang-etal-2025-lessons} to score three candidate outputs per step, retaining only the highest-reward trajectory. 
(3) \textbf{Multi-TAG Strategy} \citep{yao-yadav-2025-diverse}: An early-stopping majority vote mechanism that halts execution once the leading answer surpasses the runner-up by more than two votes. 
(4) \textbf{RaR-Gen} \citep{gunjal2026rubrics}: Replacing our retrieved rubrics with ones generated online following the RaR procedure, with the reference answer removed from the generation prompt.

\textbf{Backbone Models.}
We adopt Qwen3.5-9B \citep{qwen3.5}\footnote{https://huggingface.co/Qwen/Qwen3.5-9B} as the backbone of the Dynamic-MAS selector.
For the reasoning LLM of all participants and auditors, we deploy Qwen3-8B \cite{qwen3technicalreport} configured with the thinking mode explicitly disabled.
For the offline rubric bank construction, GPT-5.4-2026-03-05\footnote{https://developers.openai.com/api/docs/models/gpt-5.4} and GPT-4.1-mini-2025-0414 serve as the base for the miner and deduplicator, respectively.
Qwen3-Embedding-8B is adopted as the embedding model $M_\text{emb}$.

\begin{table*}[t]
    \centering
    \setlength{\aboverulesep}{0pt}
    \setlength{\belowrulesep}{0pt}
    \renewcommand{\arraystretch}{1.1} 
    \caption{Performance of our method and the baselines across mathematical benchmarks, conducted within the fixed and dynamic MAS frameworks. 
    The {\setlength{\fboxsep}{1.5pt}\colorbox{softblue!8}{blue section (left)}} represents relatively easy tasks, while the {\setlength{\fboxsep}{1.5pt}\colorbox{softorange!8}{orange section (right)}} indicates harder ones. Notably, MASRubric demonstrates the most significant performance improvements on these harder datasets. 
    ``OlymB'', ``OlymE'', and ``OlymH'' represent OlympiadBench, OlymMATH Easy, and OlymMATH Hard, respectively.}
    \label{tab:main_results}
    \scalebox{0.648}{
    \begin{tabular}{l *{5}{>{\columncolor{softblue!8}}c} *{6}{>{\columncolor{softorange!8}}c} c}
    \specialrule{\heavyrulewidth}{0pt}{0pt}
        \multirow{2.15}{*}{\textbf{Method}} & \multicolumn{5}{>{\columncolor{softblue!8}}c}{\textbf{Easy Tasks}} & \multicolumn{6}{>{\columncolor{softorange!8}}c}{\textbf{Hard Tasks}} & \multirow{2.15}{*}{\textbf{All Avg}} \\
        \cmidrule(lr){2-6} \cmidrule(lr){7-12}
        & \textbf{GSM8K} & \textbf{MATH500} & \textbf{AQuA} & \textbf{AMC23} & \textbf{Avg} & \textbf{OlymB} & \textbf{AIME24} & \textbf{AIME25} & \textbf{OlymE} & \textbf{OlymH}  & \textbf{Avg} & \\
    \specialrule{\lightrulewidth}{0pt}{0pt}
        Single Agent            & 87.64 & 74.80 & 83.86 & 62.50 & 77.20 & 47.56 & 13.33 & 20.00 & 20.00 & 16.00 & 23.38 & 47.30 \\
        \quad + CoT             & 93.71 & 79.40 & 84.65 & 67.50 & 81.32 & 43.85 & 20.00 & 23.33 & 24.00 & 15.00 & 25.24 & 50.16 \\
    \specialrule{\lightrulewidth}{0pt}{0pt}
        Fixed-MAS               & 93.18 & 77.40 & \textbf{85.04} & 65.00 & 80.15 & 49.04 & 26.67 & 20.00 & 29.00 & 15.00 & 27.94 & 51.15 \\
        \quad + Self-Refine     & 93.03 & 77.00 & 83.46 & 65.00 & 79.62 & \textbf{50.67} & 23.33 & 20.00 & 27.00 & \textbf{22.00} & 28.60 & 51.28 \\
        \quad + Multi-TAG       & 93.78 & 76.20 & 83.07 & \textbf{70.00} & 80.76 & 47.85 & 26.67 & \textbf{23.33} & 19.00 & 17.00 & 26.77 & 50.77 \\
        \quad + PRM             & \textbf{94.84} & \textbf{80.80} & 84.25 & \textbf{70.00} & \textbf{82.47} & 48.59 & 26.67 & 20.00 & 23.00 & 13.00 & 26.25 & 51.24 \\
        \quad + RaR-Gen         & 93.33 & 77.40 & 81.50 & \textbf{70.00} & 80.56 & 47.41 & 26.67 & 13.33 & 23.00 & 15.00 & 25.08 & 49.74 \\
        \quad + MASRubric       & 93.63 & 77.00 & 84.65 & 67.50 & 80.69 & 48.59 & \textbf{30.00} & \textbf{23.33} & \textbf{30.00} & 20.00 & \textbf{30.39} & \textbf{52.74} \\

    \specialrule{\lightrulewidth}{0pt}{0pt}
        Dynamic-MAS             & 91.66 & 78.00 & 85.43 & 62.50 & 79.40 & 48.15 & \textbf{30.00} & 20.00 & 26.00 & 16.00 & 28.03 & 50.86 \\
        \quad + Self-Refine     & 91.51 & 83.40 & 85.04 & 62.50 & 80.61 & 49.19 & 23.33 & 10.00 & 20.00 & 15.00 & 23.50 & 48.89 \\
        \quad + Multi-TAG       & 92.04 & 81.00 & \textbf{85.83} & 62.50 & 80.34 & 48.74 & \textbf{30.00} & 16.67 & 27.00 & \textbf{18.00} & 28.08 & 51.31 \\
        \quad + PRM             & \textbf{93.33} & \textbf{83.80} & 85.43 & 67.50 & \textbf{82.52} & 51.70 & \textbf{30.00} & 16.67 & 24.00 & 15.00 & 27.47 & 51.94 \\
        \quad + RaR-Gen         & 91.74 & 80.60 & \textbf{85.83} & \textbf{70.00} & 82.04 & 47.26 & 26.67 & 23.33 & 30.00 & 12.00 & 27.85 & 51.94 \\
        \quad + MASRubric       & 91.66 & 79.60 & 83.86 & \textbf{70.00} & 81.28 & \textbf{52.44} & \textbf{30.00} & \textbf{26.67} & \textbf{32.00} & 17.00 & \textbf{31.62} & \textbf{53.69} \\
    \specialrule{\heavyrulewidth}{0pt}{0pt}
    \end{tabular}
    }
\end{table*}

\textbf{Datasets.}
For mathematical reasoning, we employ nine benchmarks spanning a spectrum of difficulty levels, including GSM8K \citep{cobbe2021trainingverifierssolvemath}, MATH-500 \citep{ICLR2024_aca97732}, AQuA \citep{ling-etal-2017-program}, AMC23\footnote{https://huggingface.co/datasets/math-ai/amc23}, OlympiadBench \citep{he-etal-2024-olympiadbench}, OlymMATH Easy, OlymMATH Hard \citep{sun-etal-2026-challenging}, AIME24 \citep{aime24}, and AIME25 \citep{aime25}.
For code generation capabilities, we assess the model on four established datasets:
MBPP \citep{austin2021programsynthesislargelanguage}, HumanEval \citep{chen2021evaluatinglargelanguagemodels}, CodeContests \citep{doi:10.1126/science.abq1158}, and LiveCodeBench \citep{jain2025livecodebench}.
For rubric distillation, we sample trajectories to distill domain-specific pitfall rubrics using the training splits of MATH and AQuA for the mathematical domain, and the training sets of MBPP, KodCode \citep{xu-etal-2025-kodcode}, and CodeContests for the code domain.
Detailed statistics for each dataset and the rubric bank are provided in Appendix \ref{app:dataset}.

\textbf{Hyper-Parameters.}
We set the max chat turns of SelectorGroupChat to 6, and the max revision rounds $T_\text{max}$ to 3.
We set $K_\text{cand}$ to 20, $K_\text{ctx}$ to 5 for math and 3 for code during rubric retrieval, and set $K_\text{dedup}$ to 20.
The pass rate satisfaction threshold $\tau_\text{pass}$ is set to 60\% and 100\% for math and code benchmarks, respectively.
The temperature of the auditor is set to 0, and the others remain 0.7.

\subsection{Main Results}

\textbf{Enhancements in Mathematical Reasoning.}
Table \ref{tab:main_results} reports results on the mathematical benchmarks.
MASRubric achieves the highest overall average in both frameworks, raising Fixed-MAS from 51.15\% to 52.74\% and Dynamic-MAS from 50.86\% to 53.69\%.
The gains concentrate on the hard tier, where MASRubric improves the hard-task average by 2.45 and 3.59 points and leads every baseline, even though its criteria are mined only from the training splits of MATH and AQuA; on the easy tier, where the backbone already performs well, PRM remains the strongest.
Retrieving criteria from the off-the-shelf bank also beats writing rubrics on the fly: RaR-Gen, which generates a new rubric for every query, trails MASRubric in both frameworks and even falls below the vanilla Fixed-MAS (49.74\% vs.\ 51.15\%), while Self-Refine brings little benefit or degrades performance.
Multi-TAG, which only votes over final answers, brings no consistent gain, likely because a vote cannot undo an error that most agents have already adopted.

\begin{wraptable}{r}{0.5\textwidth}
    \vspace{-20pt}
    \centering
    \caption{Performance comparison of our method against baselines across code domain benchmarks. ``HumE'', ``CodeC'', ``LiveC'' represent HumanEval, CodeContests, and LiveCodeBench.}
    \label{tab:code_domain}
    \scalebox{0.653}{
    \begin{tabular}{l ccccc}
        \toprule
        \textbf{Method} & \textbf{MBPP} & \textbf{HumE} & \textbf{CodeC} & \textbf{LiveC} & \textbf{Average} \\
        \midrule
        Single Agent            & 61.87 & 85.71 & 7.27 & 29.50 & 46.09 \\
        \midrule
        Dynamic-MAS             & 65.76 & 85.09 & 6.67 & 29.00 & 46.63 \\
        \quad + Self-Refine     & 64.20 & 83.23 & 6.06 & \textbf{32.50} & 46.50 \\
        \quad + Multi-TAG       & 66.15 & 80.75 & 6.06 & 29.50 & 45.61 \\
        \quad + RaR-Gen         & 64.98 & 82.61 & 6.06 & 29.25 & 45.73 \\
        \hdashline\noalign{\vskip 0.6ex}
        \quad + \textbf{MASRubric}   & \textbf{68.48} & \textbf{85.71} & \textbf{7.27} & 32.00 & \textbf{48.37} \\
        \bottomrule
    \end{tabular}
    }
    \vspace{-6pt}
\end{wraptable}

\textbf{Enhancements in Code Generation.}
Table \ref{tab:code_domain} reports results on the code generation benchmarks.
Collaboration alone brings little in this domain: the vanilla Dynamic-MAS improves on the single agent by only 0.54 points (46.63\% vs.\ 46.09\%).
MASRubric raises the average to 48.37\%, the best in the table, with the largest gains on LiveCodeBench ($+$3.00) and MBPP ($+$2.72).
It is also the only test-time method that improves on the vanilla MAS: Self-Refine, Multi-TAG, and RaR-Gen all lower the average (46.50\%, 45.61\%, and 45.73\%).
This demonstrates that our failure-distilled rubric and audit framework are effective in the programming domain, successfully intercepting and mitigating complex logical and syntactical errors in the workflows.

\section{Analysis}

\subsection{Ablation Study}
\begin{table*}[t]
    \caption{Results of the ablation study on the Dynamic-MAS framework.}
    \label{tab:ablation}
    \centering
    \scalebox{0.653}{
    \begin{tabular}{l cccccccccc}
    \toprule
        \textbf{Method} & \textbf{GSM8K} & \textbf{MATH500} & \textbf{AQuA} & \textbf{AMC23} & \textbf{OlymB} & \textbf{AIME24} & \textbf{AIME25} & \textbf{OlymE} & \textbf{OlymH} & \textbf{Average} \\
        \midrule
        \textbf{MASRubric}  & 91.66 & 79.60 & 83.86 & 70.00 & 52.44 & 30.00 & 26.67 & 32.00 & 17.00 & 53.69 \\
        \midrule
        \multicolumn{11}{c}{\textit{(I) Max Revision Rounds ($T_\text{max}$, Default: 3)}} \\
        1 Round         & 91.28 & 81.60 & 85.43 & 65.00 & 47.70 & 33.33 & 23.33 & 24.00 & 17.00 & 52.08 \\
        2 Rounds        & 91.74 & 80.80 & 86.22 & 62.50 & 49.63 & 33.33 & 23.33 & 25.00 & 18.00 & 52.28 \\
        4 Rounds        & 90.83 & 79.60 & 82.28 & 75.00 & 48.00 & 30.00 & 20.00 & 29.00 & 11.00 & 51.75 \\
        \midrule
        \multicolumn{11}{c}{\textit{(II) Number of Retrieved Criteria ($K_\text{ctx}$, Default: 5)}} \\
        3 Criteria        & 90.83 & 81.80 & 84.65 & 62.50 & 50.52 & 30.00 & 26.67 & 30.00 & 16.00 & 52.55 \\
        7 Criteria        & 91.13 & 77.20 & 82.68 & 72.50 & 50.96 & 23.33 & 26.67 & 24.00 & 14.00 & 51.39 \\
        \midrule
        \multicolumn{11}{c}{\textit{(III) Satisfaction Threshold ($\tau_\text{pass}$, Default: 60\%)}} \\
        40\%                & 91.36 & 80.60 & 83.07 & 70.00 & 49.78 & 26.67 & 20.00 & 26.00 & 14.00 & 51.27 \\
        100\%               & 91.66 & 79.00 & 83.86 & 70.00 & 50.96 & 30.00 & 23.33 & 25.00 & 17.00 & 52.31 \\
        \midrule
        \multicolumn{11}{c}{\textit{(IV) Rubric Criterion Bank Deduplication}} \\
        w/o Deduplication   & 91.36 & 80.60 & 86.22 & 75.00 & 50.37 & 26.67 & 16.67 & 31.00 & 13.00 & 52.32 \\
        \midrule
        \multicolumn{11}{c}{\textit{(V) Criteria Selection Mechanism}} \\
        Random 1-5          & 91.05 & 81.20 & 85.04 & 70.00 & 51.85 & 26.67 & 20.00 & 24.00 & 15.00 & 51.65 \\
        Generic Criterion   & 89.92 & 81.20 & 83.86 & 70.00 & 50.81 & 36.67 & 16.67 & 26.00 & 16.00 & 52.35 \\
        \bottomrule
    \end{tabular}
    }
\end{table*}

\textbf{Impact of Max Revision Rounds.}
We first examine the impact of the revision budget $T_\text{max}$ on overall performance.
As reported in Block I of Table \ref{tab:ablation}, restricting the budget to one or two rounds (52.08\% and 52.28\%) and extending it to four (51.75\%) both fall short of the default.
Accuracy rises monotonically from one round to three, indicating that a tight budget leaves detected flaws only partially repaired, whereas the decline at four rounds suggests that excessive retries may induce over-correction or reasoning drift.
Consequently, $T_\text{max}=3$ strikes the balance between error mitigation thoroughness and system stability.

\textbf{Impact of Retrieved Criteria Count.}
Next, we explored the impact of the number of retrieved criteria as shown in Block II of Table \ref{tab:ablation}.
Both reducing the retrieved count to $K_\text{ctx}=3$ and increasing it to $K_\text{ctx}=7$ degrade performance compared to the setting of $K_\text{ctx}=5$.
This indicates that while agents benefit from diverse failure patterns, providing an excessive number of criteria results in information overload, distracting the model with less relevant constraints rather than aiding the reasoning process.

\textbf{Impact of Pass Rate Threshold.}
For the pass rate threshold $\tau_\text{pass}$ (Block III), the default 60\% achieves the highest accuracy.
Lowering it to 40\% lets more flawed messages through, whereas a strict 100\% requirement acts on any single flagged criterion and thus amplifies occasional over-eager judgments.
Both extremes still outperform the vanilla Dynamic-MAS (51.27\% and 52.31\% vs.\ 50.86\%), indicating that the mechanism is not fragile to this choice.

\textbf{Necessity of Criterion Deduplication.}
We examine the necessity of criterion deduplication in Block IV of Table \ref{tab:ablation}.
Omitting it drops the average accuracy to 52.32\%, 1.37 points below the full system.
Without deduplication, near-duplicate criteria mined from similar failures compete for the at most $K_\text{ctx}$ slots of each rubric, so a single failure mode can occupy several slots and crowd out checks for other pitfalls.
This underscores the importance of bank deduplication.

\textbf{Impact of Criterion Retrieval.}
Replacing the retrieved rubric with 1--5 criteria sampled at random lowers the average accuracy to 51.65\% (Block V of Table \ref{tab:ablation}), so the gains depend on selecting criteria that match the current reasoning situation.
A single generic criterion that checks overall logical correctness (Appendix \ref{app:prompt}) also degrades performance, dropping the average to 52.35\%, which indicates that the pitfalls mined from failures supply checks that a generic criterion cannot.

\begin{table*}[t]
    \centering
    \caption{Performance with Qwen3-14B as the backbone, using the criterion bank mined from the Qwen3-8B system without re-mining.}
    \label{tab:qwen3-14b}
    \scalebox{0.7}{
    \begin{tabular}{l cccccccccc}
    \toprule
        \textbf{Method} & \textbf{GSM8K} & \textbf{MATH500} & \textbf{AQuA} & \textbf{AMC23} & \textbf{OlymB} &  \textbf{AIME24} & \textbf{AIME25} & \textbf{OlymE} & \textbf{OlymH} &\textbf{Average} \\
    \midrule
        Single Agent            & 87.64 & 79.40 & 84.19 & 62.50 & 52.15 & 23.33 & 20.00 & 20.00 & \textbf{16.00} & 49.47 \\
        Dynamic-MAS             & 93.48 & 80.80 & 86.22 & 72.50 & 51.41 & 26.67 & 20.00 & \textbf{31.00} & 11.00 & 52.56 \\
        \quad + \textbf{MASRubric}   & \textbf{93.93} & \textbf{82.20} & \textbf{87.01} & \textbf{77.50} & \textbf{53.93} & \textbf{43.33} & \textbf{26.67} & 27.00 & 15.00 & \textbf{56.29} \\
    \bottomrule
    \end{tabular}
    }
\end{table*}

\subsection{Run-Time Behavior}

\begin{wrapfigure}{r}{0.5\textwidth}
    \vspace{-20pt}
    \centering
    \includegraphics[trim=0 6pt 0 0, clip,width=0.9\linewidth]{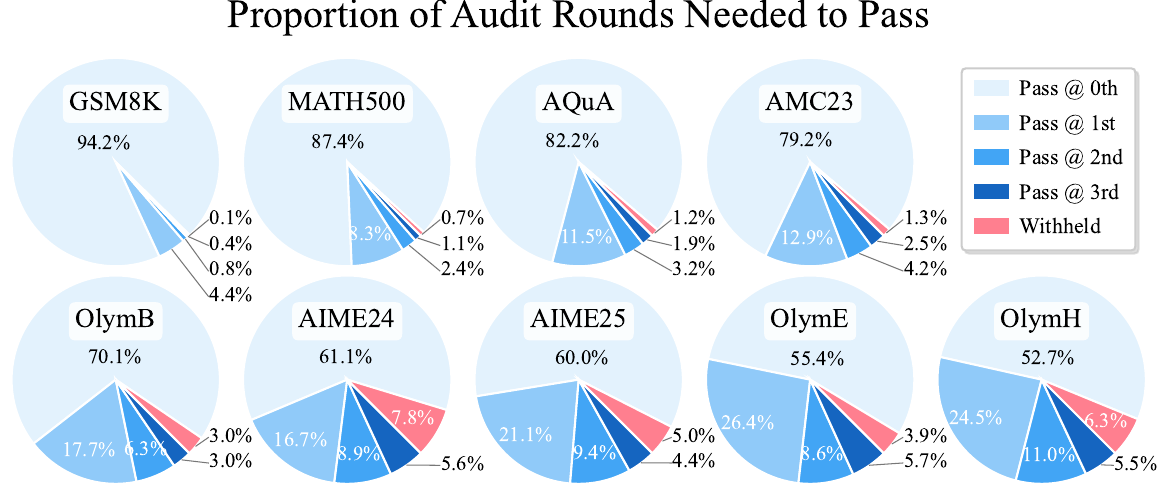}
    % \vspace{-12pt}
    \caption{Distribution of audit rounds needed to pass across benchmarks.}
    \label{fig:iteration}
    \includegraphics[width=0.74\linewidth]{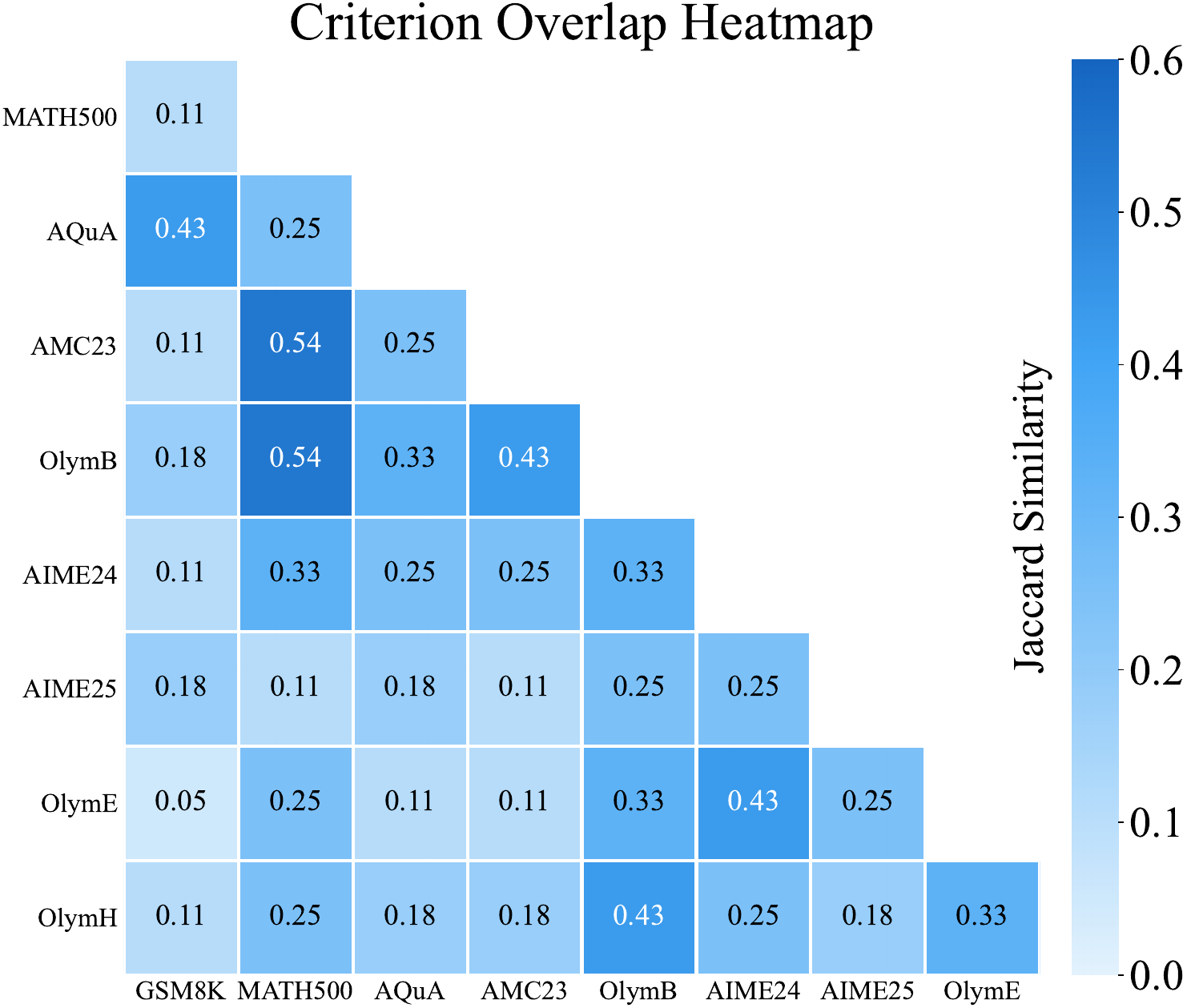}
    % \vspace{-8pt}
    \caption{Similarity between the sets of most frequently used criteria across benchmarks.}
    \label{fig:indicator_distribution}
    \vspace{-15pt}
\end{wrapfigure}

\paragraph{Audit dynamics across difficulty.}

To evaluate adaptability, we analyze iteration round distributions across varying difficulties (Figure \ref{fig:iteration}), computed over all 18{,}270 agent outputs intercepted in the Dynamic-MAS run of Table \ref{tab:main_results}; since a single problem typically involves several audited outputs, the unit of these statistics is the output rather than the problem.
``Pass @ $k$-th'' indicates outputs accepted at iteration $k$.
Revision, not withholding, is the dominant intervention: 15.4\% of intercepted outputs are accepted after revision against 1.4\% withheld, and even on the four hardest benchmarks the revised share (31--41\%) exceeds the withheld share (3.9--7.8\%) by four to ten times.
Most revised outputs pass after a single round of feedback, which accounts for the majority of revisions on every benchmark (e.g., 17.7 of the 26.9 points on OlympiadBench).
Audit depth grows with task difficulty: 94.2\% of outputs pass the first audit on GSM8K but only 52.7\% on OlymMATH Hard, where 6.3\% are ultimately withheld.
This indicates that MASRubric adapts its intervention to the task, spending little on simple queries and more where errors are intricate.
The same trend makes audit depth and withholding rate a reference-free signal of how hard a benchmark is for the backbone at hand.
Please refer to Appendix \ref{app:adaptivity} for detailed statistics.

\paragraph{Criterion usage across benchmarks.}

To assess the utility of the criterion bank, we collect for each benchmark the ten criteria retrieved most often and compute their pairwise Jaccard similarity, i.e., the size of their intersection over that of their union (Figure \ref{fig:indicator_distribution}).
The heatmap reveals distinct block-wise correlations, indicating that tasks with similar reasoning demands share common failure modes.
For instance, foundational datasets like GSM8K and AQuA share a 0.43 similarity, while harder benchmarks such as OlymMATH Easy and AIME24 also exhibit a high overlap (0.43).
Conversely, overlap drops precipitously between disparate difficulty levels, such as GSM8K and OlymMATH Easy (0.05). This confirms that error patterns are highly task-dependent, validating both the rubric bank's diverse coverage and the retrieval mechanism's precision in isolating context-aware constraints for unique domains.

\paragraph{Off-the-shelf transfer to a stronger backbone.}
\begin{wrapfigure}{r}{0.5\textwidth}
    \centering
    \includegraphics[width=0.8\linewidth]{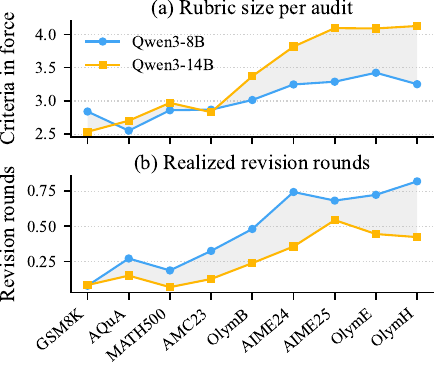}
    \vspace{-6pt}
    \caption{Run-time behavior of the auditing agent across the math suite, ordered from easy to hard. The Qwen3-14B system reuses the criterion bank mined from Qwen3-8B.
    }
    \label{fig:adaptivity}
\end{wrapfigure}

We deploy the criterion bank mined from the Qwen3-8B system directly onto a Qwen3-14B system, without re-mining and under identical caps.
As shown in Table \ref{tab:qwen3-14b}, the average accuracy rises from 52.56\% to 56.29\%, confirming that the extracted failure patterns remain applicable under a stronger backbone, where the one-off mining cost need not be paid again.
Figure \ref{fig:adaptivity} further shows that the bank is not merely still usable under the stronger backbone, but is used differently.
The average rubric sizes per message span 2.54--4.13 criteria across benchmarks at 14B against 2.55--3.42 at 8B (Figure \ref{fig:adaptivity}a): the stronger auditor puts fewer criteria in force on the easiest sets (2.54 vs.\ 2.84 on GSM8K) and substantially more on olympiad-level ones (4.13 vs.\ 3.25 on OlymMATH-Hard), allocating scrutiny by difficulty instead of applying a uniform amount of it.
At the same time, the system needs fewer revision rounds per output on every benchmark except GSM8K, where the two are effectively tied, and the gap widens with difficulty, reaching 0.42 against 0.82 average revision rounds on OlymMATH-Hard (Figure \ref{fig:adaptivity}b).
The stronger system therefore makes more adaptive and more efficient use of the same bank, targeting its scrutiny more sharply while spending less effort on repair, suggesting that MASRubric stands to benefit from stronger backbones.
Please refer to Appendix \ref{app:adaptivity} for full statistics.

\paragraph{Accuracy-cost trade-off.}
As a test-time method, MASRubric trades inference compute for accuracy: on the Dynamic-MAS math suite it consumes 201.22M tokens against 76.43M for vanilla AutoGen, a 2.63$\times$ budget that returns $+$2.83 accuracy points.
Building the criterion bank is a one-off offline cost, whose API tokens amount to only 6.3\% of a single evaluation pass.
Comparisons with other test-time methods and the full construction figures are reported in Appendix \ref{app:cost}.

\section{Conclusion}

In this paper, we introduced MASRubric, an auditing framework that distills pitfall rubrics from a MAS's own failure trajectories and applies them at test time to audit and intervene in its information flow.
Offline, failed trajectories are automatically distilled into an off-the-shelf bank of pitfall criteria with applicability conditions; online, the criteria that match each intermediate message are retrieved, checked one by one, and turned into a decision to broadcast, revise, or withhold the message.
Experiments on fixed and dynamic MAS show consistent gains in mathematical reasoning and code generation, with the largest improvements on olympiad-level problems.
Further analysis reveals two properties of the bank at run time: the criteria it supplies vary with task type and difficulty, and the audit effort they induce tracks task difficulty.
The bank also transfers without re-mining to a system with a stronger backbone, which makes more adaptive and more efficient use of it.

\bibliography{iclr2027_conference}
\bibliographystyle{iclr2027_conference}

\appendix
\section{Appendix}

\subsection{Pseudo Codes}\label{app:codes}

The pseudo-code for the offline failure-driven rubric construction is outlined in Algorithm \ref{alg:bank_construction}.
The process begins by iterating through the source dataset to collect execution trajectories where the MAS fails to deliver the correct solution (Lines 3-4).
Subsequently, the rubric miner scrutinizes these failure instances against the reference answer, synthesizing candidate criteria that capture the specific error patterns exhibited by individual agents (Lines 5-6).
To keep the bank compact, a redundancy elimination mechanism is applied to each candidate.
The algorithm first encodes the new criterion into a semantic vector to retrieve the most similar existing entries (Lines 7-9). A deduplication model then verifies the novelty of the candidate, admitting it into the bank only if it represents a distinct and previously unrecorded error pattern (Lines 10-12).

Algorithm \ref{alg:intervention} outlines the pseudo-code for test-time rubric auditing and intervention.
During MAS execution, the output of each agent is intercepted before it is broadcast to its successors.
First, a candidate criterion set is retrieved from the bank by semantic matching against the applicability conditions, and the auditing agent then selects the context-specific rubric from this set (Lines 5-8).
The auditing agent subsequently judges the output against each selected criterion, emitting a diagnostic rationale as feedback whenever a criterion is violated (Lines 10-13).
The per-criterion judgments are aggregated into the rubric satisfaction rate, which drives a tri-state gating decision: the iteration terminates once the satisfaction rate meets the satisfaction threshold, or once the revision budget is exhausted (Lines 14-23).
Finally, an accepted message is propagated to successor agents (Lines 24-26), while a message that never reaches the threshold is withheld.

\subsection{Criterion \& Prompt Design}\label{app:prompt}

\textbf{Criterion Design.}
Figure \ref{fig:indicator_example} displays a criterion from our constructed bank, with all four fields of Equation \ref{eq:criterion} instantiated.
The division of labor across fields is visible here: the diagnostic definition names the misconception behind the failure, the applicability condition states when percentage-coverage reasoning is at stake, and the inspection directive specifies the quantity the auditing agent has to re-examine.
We also design a single generic criterion for a comparison, as illustrated in Figure \ref{fig:indicator_math}.
It preserves the same four-field structure, but its applicability condition matches the whole domain, so no routing takes place.

\textbf{Prompt Design.}
The prompt templates of the auditing agent for the math and code domains are presented in Figure \ref{fig:prompt_math} and Figure \ref{fig:prompt_code}.
Additionally, Figure \ref{fig:prompt_teacher} depicts the prompt template for the rubric miner, which generates new criteria from failed MAS execution trajectories.

\subsection{Case Study}\label{app:case}

\begin{figure*}
    \centering
\begin{tcolorbox}[notitle, colback=white, colframe=softpurple,
       fontupper=\small,
       boxrule=3pt, boxsep=0.5pt, enhanced, 
       shadow={3pt}{-3pt}{0pt}{opacity=1,mygray},
       title={\large \includegraphics[height=22pt]{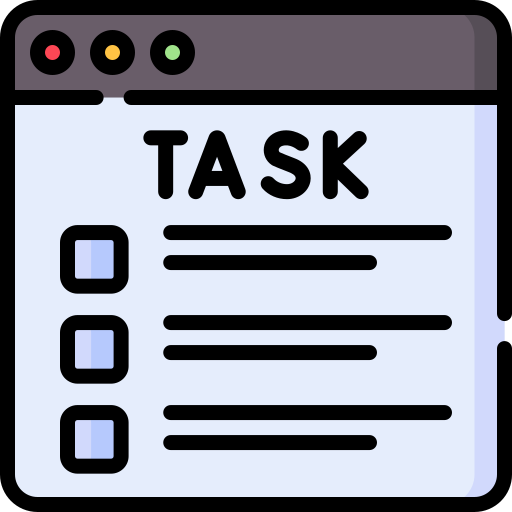} {\large Math Task $\mathcal{Q}$ and Correct Answer $\mathcal{Y^*}$}}]
Task:\\
Let \(S\) be the union of the set of all points inside a regular nonagon with
side length \(2\) units and the set of all points less than \(1\) unit away
from a point on the perimeter of the nonagon. What, in units, is the perimeter
of \(S\)?\\

Correct Answer:\\
\(\boxed{18+2\pi}\)\\

Reference logic:\\
The offset boundary keeps nine straight portions of length \(2\), contributing
\(18\). The rounded pieces around the nine vertices have total turning angle
\(2\pi\), and radius \(1\), contributing \(2\pi\). Hence the perimeter is
\(18+2\pi\).

\end{tcolorbox}
    \caption{An example of the given math task.}
    \label{fig:case_task}
\end{figure*}

\begin{figure*}
    \centering
\begin{tcolorbox}[notitle, colback=white, colframe=softblue,
       fontupper=\small,
       boxrule=3pt, boxsep=0.5pt, enhanced, 
       shadow={3pt}{-3pt}{0pt}{opacity=1,mygray},
       title={\large \includegraphics[height=22pt]{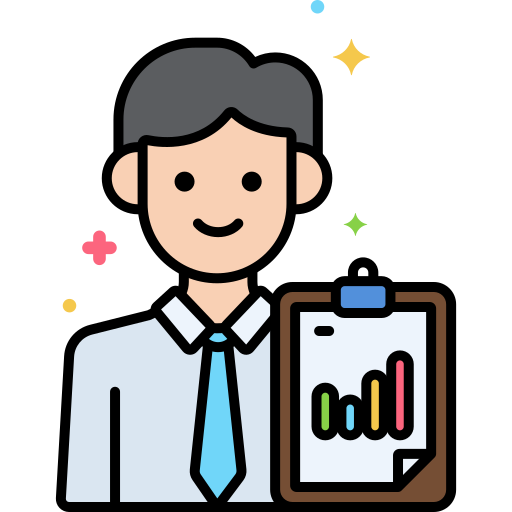} {\large Math Solver: Initial Output $o^{(0)}$}}]
The solver is asked to verify two candidate answers for the nonagon buffer
perimeter:\\

- Participant\_1: \(\boxed{18+\frac{9}{2}\pi}\), from treating the nine rounded
  corners as quarter-circles.\\
- Participant\_2: \(\boxed{18+7\pi}\), from using the regular nonagon's internal
  angle for the nine rounded-corner arcs.\\

It first computes the original nonagon perimeter:

\[
9 \times 2 = 18.
\]

It then rejects the quarter-circle interpretation and accepts the internal-angle
interpretation. The solver states that the internal angle of a regular nonagon
is

\[
\frac{(9-2)\pi}{9}=\frac{7\pi}{9}.
\]

Using radius \(1\), it calculates each corner arc as

\[
1\cdot\frac{7\pi}{9}=\frac{7\pi}{9},
\]

so the nine arcs contribute

\[
9\cdot\frac{7\pi}{9}=7\pi.
\]

The straight portions contribute \(18\). The solver therefore outputs

\[
\boxed{18+7\pi}.
\]
\end{tcolorbox}
    \caption{The initial output of the math solver agent.}
    \label{fig:case_reasoner_0}
\end{figure*}

\begin{figure*}
    \centering
\begin{tcolorbox}[notitle, colback=white, colframe=softorange,
       fontupper=\small,
       boxrule=3pt, boxsep=0.5pt, enhanced, 
       shadow={3pt}{-3pt}{0pt}{opacity=1,mygray},
       title={\large \includegraphics[height=22pt]{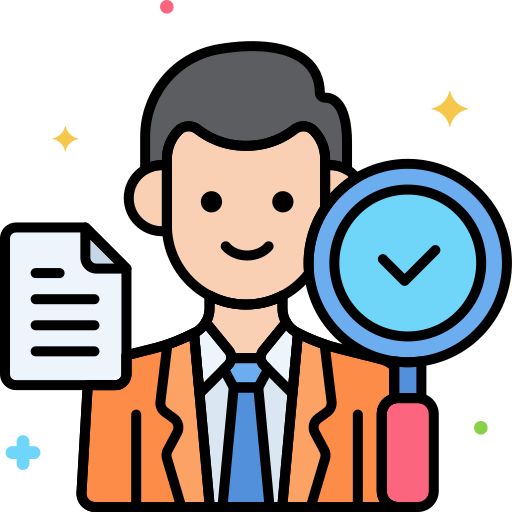} {\large Auditor: Revision Round 0/2 Evaluation}}]
[1] Matched Criteria:

\quad - PERIMETER\_CALCULATION\_MISCOMPONENTS

\quad - PERIMETER\_PROPAGATION\_ERROR

\quad - GEOMETRIC\_COMPOUND\_CALCULATION\_CHECK

\quad - INCORRECT\_PERIMETER\_FORMULA

\quad - ANGLE\_DEPENDENCY\_CHECK\\

[2] Evaluations:

\quad - PERIMETER\_CALCULATION\_MISCOMPONENTS: Flawed

  \quad \quad Evidence: Participant\_1 claims \(18+\frac{9}{2}\pi\).
  
  \quad \quad Analysis: The quarter-circle interpretation uses a fixed \(90^\circ\) arc
  at each vertex, which does not match the geometry of a regular nonagon.
  
  \quad \quad Suggestion: Replace the quarter-circle assumption with arcs tied to the
  nonagon's angle.\\

\quad - PERIMETER\_PROPAGATION\_ERROR: Flawed

  \quad \quad Evidence: The solver discusses rounded corners and the propagation from local
  corner shape to total perimeter.
  
  \quad \quad Analysis: The local arc angle is propagated into the final perimeter formula.
  
  \quad \quad Suggestion: Re-evaluate the angle used for every rounded corner.\\

\quad - GEOMETRIC\_COMPOUND\_CALCULATION\_CHECK: Flawed

  \quad \quad Evidence: The response compares a quarter-circle calculation with an
  internal-angle calculation.
  
  \quad \quad Analysis: The calculation combines straight side contributions and circular
  arc contributions, but the corner-angle component is under-audited.
  
  \quad \quad Suggestion: Recompute the corner sector before summing with the straight
  contribution.\\

\quad - INCORRECT\_PERIMETER\_FORMULA: Flawed

  \quad \quad Evidence: The final formula adds \(18\) and \(7\pi\).
  
  \quad \quad Analysis: The formula inherits the chosen arc angle directly, so an incorrect
  angle selection changes the final perimeter.
  
  \quad \quad Suggestion: Rebuild the perimeter formula from the offset-boundary geometry.\\

\quad - ANGLE\_DEPENDENCY\_CHECK: Flawed

  \quad \quad Evidence: The solver chooses between \(90^\circ\) and \(7\pi/9\).
  
  \quad \quad Analysis: The answer depends on choosing the correct angle type for the arc.
  
  \quad \quad Suggestion: Audit whether the arc angle should follow the polygon's internal
  angle, external angle, or another derived angle.\\

[3] Pass Count:

0/5, threshold 60\%, action: revise.\\

[4] Feedback Sent to Solver:

The auditor flags the quarter-circle assumption and asks the solver to revise
the corner arcs using the regular nonagon's angle. The feedback still points the
solver toward the internal-angle explanation, which leaves room for a second
wrong answer.
\end{tcolorbox}
    \caption{The auditor's judgments and feedback to the initial output.}
    \label{fig:case_rectifier_0}
\end{figure*}

\begin{figure*}
    \centering
\begin{tcolorbox}[notitle, colback=white, colframe=softblue,
       fontupper=\small,
       boxrule=3pt, boxsep=0.5pt, enhanced, 
       shadow={3pt}{-3pt}{0pt}{opacity=1,mygray},
       title={\large \includegraphics[height=22pt]{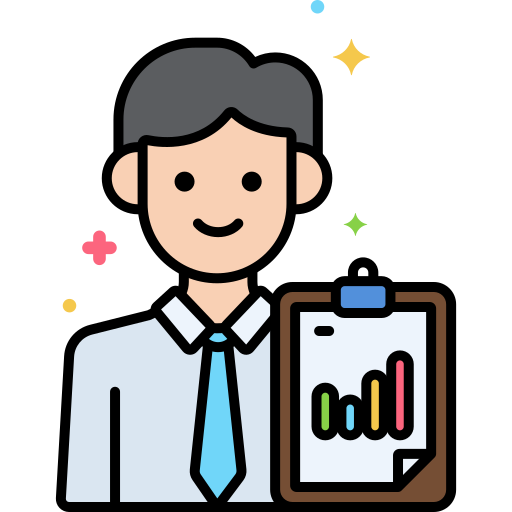} {\large Math Solver: Revision Round 1/2 Output $o^{(1)}$}}]
The solver accepts the first audit's feedback. It states that the original
quarter-circle assumption was invalid because the internal angle of a regular
nonagon is

\[
\frac{7\pi}{9},
\]

and then treats this internal angle as the angle of each rounded corner.\\

The solver describes \(S\) as a Minkowski-style offset of the nonagon by a disk
of radius \(1\). It then decomposes the boundary into straight portions and
rounded corners:\\

1. Internal angle of the nonagon:

\[
\frac{(9-2)\pi}{9}=\frac{7\pi}{9}.
\]

2. Arc length for each rounded corner:

\[
1\cdot\frac{7\pi}{9}=\frac{7\pi}{9}.
\]

3. Total rounded-corner contribution:

\[
9\cdot\frac{7\pi}{9}=7\pi.
\]

4. Straight-side contribution:

\[
9\times 2=18.
\]

The solver therefore keeps the same final answer:

\[
\boxed{18+7\pi}.
\]
\end{tcolorbox}
    \caption{The output of the math solver agent in Revision Round 1.}
    \label{fig:case_reasoner_1}
\end{figure*}

\begin{figure*}
    \centering
\begin{tcolorbox}[notitle, colback=white, colframe=softorange,
       fontupper=\small,
       boxrule=3pt, boxsep=0.5pt, enhanced, 
       shadow={3pt}{-3pt}{0pt}{opacity=1,mygray},
       title={\large \includegraphics[height=22pt]{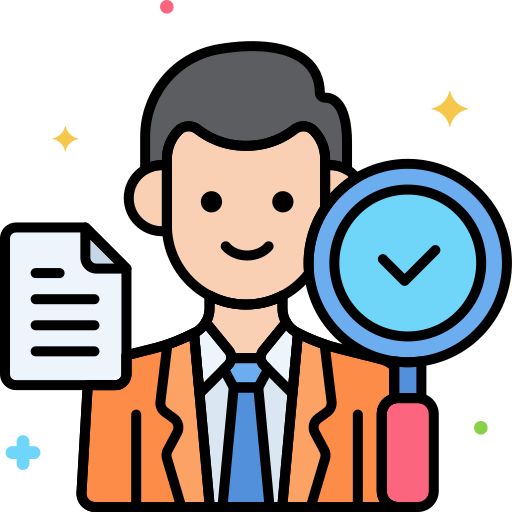} {\large Auditor: Revision Round 1/2 Evaluation}}]
[1] Matched Criteria:

\quad - ANGLE\_DEPENDENCY\_CHECK

\quad - GEOMETRIC\_COMPOUND\_CALCULATION\_CHECK

\quad - CALCULATION\_CONTINGENT\_ON\_ANGLES

\quad - PERIMETER\_PROPAGATION\_ERROR

\quad - IGNORING\_STRAIGHT\_SEGMENT\_CONTRIBUTIONS\\

[2] Evaluations:

\quad - ANGLE\_DEPENDENCY\_CHECK: Correct

  The response explicitly recognizes that the perimeter calculation depends on
  the angle assigned to each rounded corner.

\quad - GEOMETRIC\_COMPOUND\_CALCULATION\_CHECK: Flawed

  \quad \quad Evidence: The response uses the nonagon's internal angle \(7\pi/9\) as the
  rounded-corner arc angle.
  
  \quad \quad Analysis: For a unit offset of a polygon, the rounded corner is determined by
  the exterior turning angle. For a regular nonagon, this angle is \(2\pi/9\).
  
  \quad \quad Suggestion: Use the external angle \(2\pi/9\) for each radius-1 arc.
  Impact: The arc contribution changes from \(7\pi\) to \(2\pi\), so the final
  perimeter changes from \(18+7\pi\) to \(18+2\pi\).\\

\quad - CALCULATION\_CONTINGENT\_ON\_ANGLES: Flawed

  \quad \quad Evidence: The solver computes \(1\cdot 7\pi/9\) per corner.
  
  \quad \quad Analysis: The arithmetic follows from the chosen angle, but the chosen angle
  type is wrong. The offset boundary uses the external angle.
  
  \quad \quad Suggestion: Compute each arc as
  \(1\cdot 2\pi/9=2\pi/9\), giving total arc length \(2\pi\).\\

\quad - PERIMETER\_PROPAGATION\_ERROR: Flawed

  \quad \quad Evidence: The final answer \(18+7\pi\) is produced by propagating the
  \quad \quad internal-angle arc length across nine corners.
  
  Analysis: The wrong angle is carried through the perimeter calculation.
  
  \quad \quad Suggestion: Replace the total arc term \(7\pi\) with \(2\pi\).\\

\quad - IGNORING\_STRAIGHT\_SEGMENT\_CONTRIBUTIONS: Correct

  The solver retains the straight-side contribution \(9\times 2=18\).\\

[3] Pass Count:

2/5, threshold 60\%, action: revise.\\

[4] Feedback Sent to Solver:

An external auditor has reviewed the previous output and flagged the core angle
type error. The feedback states that the rounded-corner arc is controlled by the
regular nonagon's external angle, \(2\pi/9\), rather than its internal angle,
\(7\pi/9\). With radius \(1\), each arc has length \(2\pi/9\), and the nine
arcs contribute \(2\pi\). The corrected target expression is therefore
\(18+2\pi\).
\end{tcolorbox}
    \caption{The auditor's judgments and feedback to the Round 1 output.}
    \label{fig:case_rectifier_1}
\end{figure*}

This case study exemplifies the framework's capability to navigate complex constraint satisfaction problems through a rubric-aware audit process.
The agent was tasked with the math problem shown in Figure \ref{fig:case_task}.
The initial solver output (Figure \ref{fig:case_reasoner_0}) and the first revision (Figure \ref{fig:case_reasoner_1}) both end at the same wrong answer, \(18+7\pi\), because both use the internal angle of the regular nonagon for the offset-boundary arcs.
The first audit (Figure \ref{fig:case_rectifier_0}) flags the response but only moves the solver away from a quarter-circle assumption; the second audit (Figure \ref{fig:case_rectifier_1}) then isolates the decisive angle-type error and directs the solver to the external angle \(2\pi/9\).
After applying that feedback, the solver changes the arc contribution from \(7\pi\) to \(2\pi\), produces \(\boxed{18+2\pi}\), and passes all five selected criteria.
This trajectory demonstrates the effectiveness of the rubric-aware audit mechanism in stabilizing reasoning through iterative, multi-dimensional constraint enforcement.

\begin{table*}[t]
    \centering
    \caption{Mean and distribution of the criterion number actually put in force per audit. Both systems use the same criterion bank, mined from the Qwen3-8B system.}
    \label{tab:rubric_size}
    \scalebox{0.7}{
    \begin{tabular}{l c cccccc c cccccc}
    \toprule
        & \multicolumn{7}{c}{\textbf{Qwen3-8B}} & \multicolumn{7}{c}{\textbf{Qwen3-14B}} \\
        \cmidrule(lr){2-8} \cmidrule(lr){9-15}
        \textbf{Benchmark} & \textbf{Mean} & \textbf{0} & \textbf{1} & \textbf{2} & \textbf{3} & \textbf{4} & \textbf{5}
                           & \textbf{Mean} & \textbf{0} & \textbf{1} & \textbf{2} & \textbf{3} & \textbf{4} & \textbf{5} \\
    \midrule
        GSM8K & 2.840 & 0.3 & 2.5 & 24.0 & 62.6 & 7.4 & 3.3 & 2.535 & 5.7 & 5.9 & 35.9 & 40.3 & 5.7 & 6.4 \\
        MATH500 & 2.859 & 0.8 & 4.1 & 24.9 & 56.5 & 6.0 & 7.7 & 2.970 & 12.8 & 2.9 & 14.9 & 36.3 & 10.0 & 23.1 \\
        AQuA & 2.552 & 1.7 & 6.7 & 37.7 & 45.6 & 4.9 & 3.3 & 2.702 & 14.3 & 4.3 & 20.4 & 35.8 & 8.1 & 17.1 \\
        AMC23 & 2.868 & 0.3 & 2.3 & 22.7 & 64.5 & 5.3 & 4.9 & 2.830 & 19.6 & 2.2 & 11.5 & 33.3 & 8.9 & 24.4 \\
        OlymB & 3.013 & 0.8 & 2.5 & 21.5 & 56.7 & 6.9 & 11.7 & 3.371 & 10.1 & 1.9 & 10.4 & 31.1 & 11.3 & 35.2 \\
        AIME24 & 3.247 & 0.0 & 0.3 & 14.1 & 62.2 & 7.4 & 16.0 & 3.816 & 10.7 & 0.8 & 3.7 & 20.5 & 9.8 & 54.5 \\
        AIME25 & 3.288 & 0.0 & 2.3 & 12.9 & 57.0 & 9.3 & 18.5 & 4.094 & 4.3 & 0.4 & 5.0 & 18.3 & 15.8 & 56.1 \\
        OlymE & 3.423 & 0.1 & 1.2 & 13.3 & 51.7 & 9.0 & 24.7 & 4.089 & 4.8 & 0.6 & 3.8 & 19.6 & 14.0 & 57.2 \\
        OlymH & 3.251 & 0.4 & 2.2 & 16.0 & 54.1 & 7.8 & 19.5 & 4.126 & 3.5 & 1.1 & 4.9 & 20.6 & 9.6 & 60.3 \\
    \midrule
        \textbf{All} & \textbf{2.921} & \textbf{0.6} & \textbf{3.0} & \textbf{23.5} & \textbf{57.8} & \textbf{6.9} & \textbf{8.2} & \textbf{2.982} & \textbf{8.7} & \textbf{3.8} & \textbf{21.7} & \textbf{34.9} & \textbf{8.6} & \textbf{22.4} \\
    \bottomrule
    \end{tabular}
    }
\end{table*}

\begin{table}[t]
    \centering
    \caption{Average revision rounds actually consumed per message. Column 0 denotes the acceptance at the first audit; the last column merges messages accepted after three repairs with those ultimately withheld.}
    \label{tab:depth}
    \scalebox{0.7}{
    \begin{tabular}{l c cccc c cccc}
    \toprule
        & \multicolumn{5}{c}{\textbf{Qwen3-8B}} & \multicolumn{5}{c}{\textbf{Qwen3-14B}} \\
        \cmidrule(lr){2-6} \cmidrule(lr){7-11}
        \textbf{Benchmark} & \textbf{Mean} & \textbf{0} & \textbf{1} & \textbf{2} & \textbf{3}
                           & \textbf{Mean} & \textbf{0} & \textbf{1} & \textbf{2} & \textbf{3} \\
    \midrule
        GSM8K & 0.077 & 94.2 & 4.4 & 0.8 & 0.5 & 0.082 & 92.4 & 7.2 & 0.3 & 0.1 \\
        MATH500 & 0.186 & 87.4 & 8.3 & 2.4 & 1.8 & 0.067 & 94.4 & 4.8 & 0.6 & 0.2 \\
        AQuA & 0.271 & 82.2 & 11.5 & 3.2 & 3.1 & 0.149 & 87.4 & 10.8 & 1.4 & 0.5 \\
        AMC23 & 0.325 & 79.2 & 12.9 & 4.2 & 3.8 & 0.125 & 90.0 & 7.9 & 1.7 & 0.4 \\
        OlymB & 0.481 & 70.1 & 17.7 & 6.3 & 6.0 & 0.238 & 81.1 & 15.5 & 2.0 & 1.4 \\
        AIME24 & 0.744 & 61.1 & 16.7 & 8.9 & 13.3 & 0.356 & 76.7 & 15.0 & 4.4 & 3.9 \\
        AIME25 & 0.683 & 60.0 & 21.1 & 9.4 & 9.4 & 0.544 & 66.1 & 21.7 & 3.9 & 8.3 \\
        OlymE & 0.724 & 55.4 & 26.4 & 8.6 & 9.6 & 0.445 & 65.3 & 27.0 & 5.5 & 2.2 \\
        OlymH & 0.820 & 52.7 & 24.5 & 11.0 & 11.8 & 0.423 & 67.7 & 24.3 & 6.0 & 2.0 \\
    \midrule
        \textbf{All} & \textbf{0.262} & \textbf{83.3} & \textbf{10.4} & \textbf{3.3} & \textbf{3.1} & \textbf{0.151} & \textbf{87.6} & \textbf{10.4} & \textbf{1.3} & \textbf{0.7} \\
    \bottomrule
    \end{tabular}
    }
\end{table}

\begin{table}[t]
    \centering
    \caption{End-to-end token accounting on the Dynamic-MAS math suite. ``Generic Criterion'' replaces the retrieved rubric with the single domain-wide criterion of Appendix \ref{app:prompt}.}
    \label{tab:cost}
    \scalebox{0.7}{
    \setlength{\tabcolsep}{5pt}
    \begin{tabular}{lcccccc}
    \toprule
        \textbf{Method} & \textbf{Math Avg} & \textbf{Total tok.} & \textbf{Tok./inst.} & \textbf{Cost} & \textbf{$\Delta$Acc.} \\
        & & ($\times 10^6$) & ($\times 10^3$) & \textbf{vs.\ vanilla} & \\
    \midrule
        Dynamic-MAS                     & 50.86 & 76.43  & 25.1 & 1.00$\times$ & --- \\
        \quad + Multi-TAG               & 51.31 & 69.71  & 22.9 & 0.91$\times$ & +0.45 \\
        \quad + PRM                     & 51.94 & 159.07 & 52.2 & 2.08$\times$ & +1.08 \\
        \quad + RaR-Gen                 & 51.94 & 144.85 & 47.5 & 1.90$\times$ & +1.08 \\
        \quad + Self-Refine             & 48.89 & 211.07 & 69.2 & 2.76$\times$ & $-$1.97 \\
    \midrule
        \quad + Generic Criterion) & 52.35 & 161.69 & 53.0 & 2.12$\times$ & +1.49 \\
        \quad + MASRubric          & \textbf{53.69} & 201.22 & 66.0 & 2.63$\times$ & \textbf{+2.83} \\
    \bottomrule
    \end{tabular}}
\end{table}

\begin{figure}[t]
    \centering
    \includegraphics[width=0.5\linewidth]{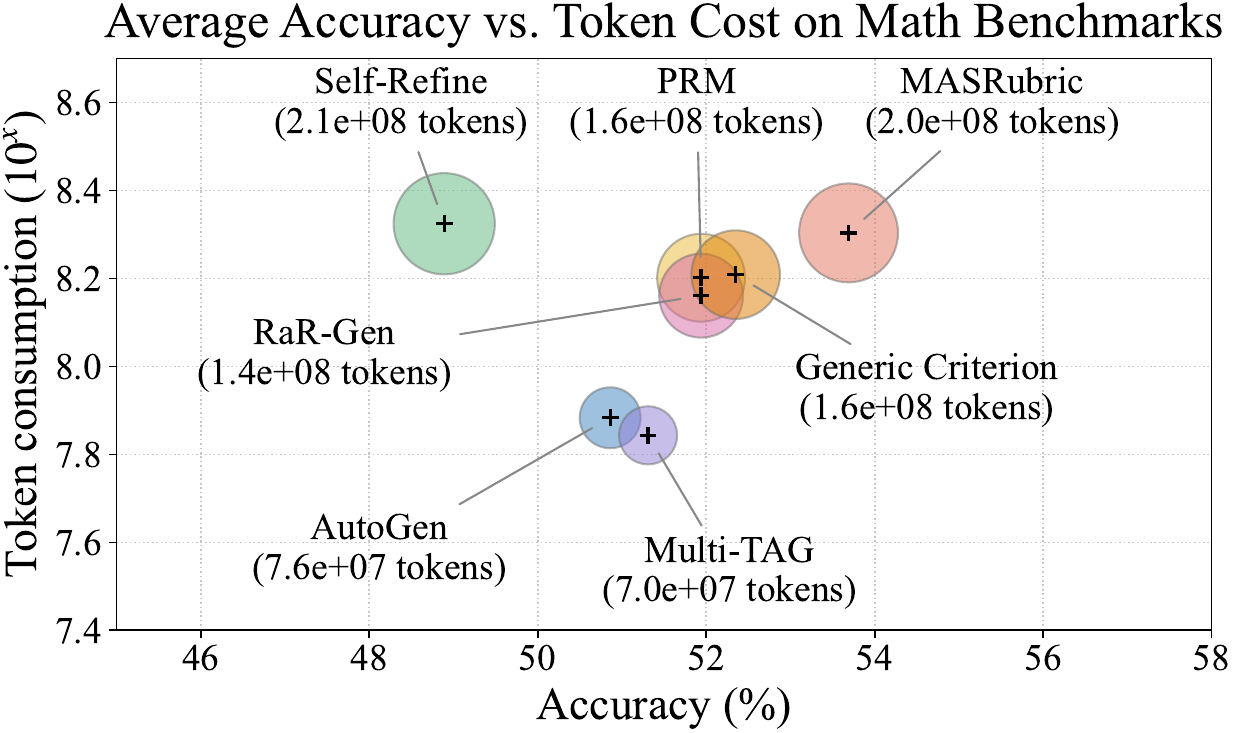}
    \caption{Average accuracy versus token consumption on Math benchmarks by the dynamic MAS framework.}
    \label{fig:performance_vs_cost}
\end{figure}

\subsection{Dataset Statistics}\label{app:dataset}

Table \ref{tab:statistics} lists the detailed statistics of the size of the datasets and the constructed rubric criterion bank.
The rubric bank for the math domain is constructed on the failed MAS trajectories on the sampled instances from the MATH and AQuA training sets, while the bank for the code domain is derived from similar failed trajectories on the MBPP, KodCode, and CodeContests training sets.

\subsection{Run-Time behavior of the Auditing Agent}\label{app:adaptivity}

Tables \ref{tab:rubric_size} and \ref{tab:depth} give the distributions summarized in Figure \ref{fig:adaptivity}.
Both systems run under the same caps and the same criterion bank; only the backbone differs.
The reduction in mean audit depth under the stronger backbone comes from fewer deep repairs rather than from a uniform shift.
The 14B system accepts 87.6\% of messages at the first audit against 83.3\% at 8B, and needs two or more repair rounds on 2.0\% of messages against 6.4\%.

\subsection{Token Accounting}\label{app:cost}

Table \ref{tab:cost} reports the complete token accounting on the Dynamic-MAS math suite, and Figure \ref{fig:performance_vs_cost} plots accuracy against token consumption for the same runs.

\textbf{Where the tokens go.}
Of the 201.22M tokens consumed by the full system, the auditing agent accounts for 102.19M (50.8\%), covering keyword extraction, criterion selection, and the per-criterion judgments.
The remaining 99.03M is spent by the reasoning agents themselves, up from 76.43M under vanilla AutoGen; the increase reflects regenerated outputs and the diagnostic feedback appended to their context.
Retrieval is negligible by comparison: the embedding model adds 0.483M input tokens, under 0.3\% of the total.

\begin{wraptable}{r}{0.5\textwidth}
    \vspace{-8pt}
    \centering
    \caption{Dataset statistics}
    \label{tab:statistics}
    \scalebox{0.7}{
    \begin{tabular}{llc}
        \toprule
        Domain & Dataset & Size \\
        \midrule
        \multicolumn{3}{c}{\textit{Test Set}} \\
        \hdashline\noalign{\vskip 0.6ex}
        \multirow{9}{*}{Math}
         & GSM8K \citep{cobbe2021trainingverifierssolvemath}                 & 1,319\\
         & MATH-500 \cite{ICLR2024_aca97732}              & 500 \\
         & AQuA \citep{ling-etal-2017-program}             & 254 \\
         & AMC23           & 40 \\
         & OlympiadBench \citep{he-etal-2024-olympiadbench}    & 675 \\
         & AIME24 \citep{aime24}                        & 30 \\
         & AIME25 \citep{aime25}                        & 30 \\
         & OlymMATH Easy \citep{sun-etal-2026-challenging} & 100 \\
         & OlymMATH Hard \citep{sun-etal-2026-challenging} & 100 \\
        \hdashline\noalign{\vskip 0.6ex}
        \multirow{4}{*}{Code}
         & MBPP \citep{austin2021programsynthesislargelanguage}               & 257 \\
         & HumanEval \citep{chen2021evaluatinglargelanguagemodels}              & 161 \\
         & CodeContests \citep{doi:10.1126/science.abq1158} & 165 \\
         & LiveCodeBench \citep{jain2025livecodebench}    & 400 \\
        \midrule
        \multicolumn{3}{c}{\textit{Training Set}} \\
        \hdashline\noalign{\vskip 0.6ex}
        \multirow{2}{*}{Math}
         & MATH            & 2,000 \\
         & AQuA            & 2,000 \\
        \hdashline\noalign{\vskip 0.6ex}
        \multirow{3}{*}{Code}
         & MBPP                            & 120 \\
         & KodCode \citep{xu-etal-2025-kodcode}   & 10,000 \\
         & CodeContests                    & 1,000 \\
        \midrule
        \multicolumn{3}{c}{\textit{Rubric Criterion Bank}} \\
        \hdashline\noalign{\vskip 0.6ex}
        Math & - & 2,000 \\
        Code & - & 2,545 \\
        \bottomrule
    \end{tabular}}
    \vspace{-8pt}
\end{wraptable}

\textbf{Spending more does not by itself buy accuracy.}
Self-Refine is the most expensive configuration in the table (2.76$\times$ vanilla) and is the only one that degrades performance ($-$1.97), leaving it worse than every alternative on both axes.
Prompting an agent to revise its own output evidently consumes a comparable budget to auditing it against explicit criteria, but converts that budget into revisions that do not survive as accuracy.

\textbf{Comparison at a matched budget.}
The generic-criterion variant consumes essentially the same budget as PRM (161.69M vs.\ 159.07M, $+$1.6\%) while reaching a slightly higher average (52.35 vs.\ 51.94); the full bank adds a further $+$1.75 points over PRM for $+$26\% tokens.
Measured against vanilla, each accuracy point costs 44.1M additional tokens under the full system and 76.5M under PRM, and the bank's own increment over the generic criterion is the cheapest segment of the curve at 29.5M tokens per point ($+$39.53M tokens for $+$1.34 points).

\textbf{The bank is built once, offline.}
The figures above concern inference alone.
Constructing the math criterion bank required roughly 4K problem-level rollouts of the Qwen3-8B MAS to collect failure trajectories, consuming 97.13M tokens on local GPUs---this figure covers all rollouts, including the successful ones that yield no criteria---followed by 5.90M miner tokens for mining and 6.79M tokens for deduplication, both served through commercial APIs, yielding 2{,}000 retained criteria.
Two properties make this a one-off cost rather than a recurring one.
First, neither API model appears anywhere in the inference loop: at test time the system runs only the local backbone and the embedding model, so the 12.69M API tokens are never re-spent, however many queries are subsequently evaluated.
For scale, that entire API budget amounts to 6.3\% of the tokens a single evaluation pass of the deployed system consumes, and the full construction cost including local rollouts amounts to 54.6\%.
Second, the resulting bank is transferable: the same criteria mined from a Qwen3-8B system are applied unchanged to a Qwen3-14B system in Table \ref{tab:qwen3-14b}, so a stronger backbone does not require re-mining.

\begin{algorithm2e*}[t]
\caption{Offline Procedure: Failure-Driven Rubric Construction}
\label{alg:bank_construction}
\SetKwInOut{Input}{Input}
\SetKwInOut{Output}{Output}
\SetKwInOut{Param}{Parameters}

\Input{Source Dataset $\mathcal{D}_\text{src}=\{\mathcal{Q}, \mathcal{Y}^*\}$, Rubric miner $\Phi_\text{miner}$, Deduplication Model $\Phi_\text{dedup}$, Embedding Model $M_\text{emb}$}
\Param{Retrieval size $K_\text{dedup}$}
\Output{Rubric criterion bank $\mathcal{B}$}

\BlankLine
\tcp{Initialize empty criterion bank}
$\mathcal{B} \leftarrow \emptyset$\;

\ForEach{Instance $(\mathcal{Q}, \mathcal{Y}^*) \in \mathcal{D}_\text{src}$}{
    \tcp{Step 1: Failure Trajectory Collection}
    Execute MAS to obtain trajectory: $\mathcal{T} = (\mathcal{Q}, A_{1:N}, o_{1:N}, \mathcal{Y})$\;
    
    \If{$\mathcal{Y} \neq \mathcal{Y}^*$}{
        \tcp{Step 2: Offline Rubric Mining}
        \ForEach{Agent $A_i$ in $\mathcal{T}$}{
            $\mathcal{B}_\text{new}^{(i)} \leftarrow \Phi_\text{miner}\left(\mathcal{T}, \mathcal{Y}^*, \mathcal{R}_i, o_i \right)$\ \tcp*{Generate $m_i \ge 0$ candidates}
            
            \tcp{Step 3: Redundancy Elimination}
            \ForEach{Criterion $C_\text{new} = (n_\text{new}, d_\text{new}, a_\text{new}, s_\text{new}) \in \mathcal{B}_\text{new}^{(i)}$}{
                $\mathbf{v}_\text{new} \leftarrow M_\text{emb}(d_\text{new} \oplus a_\text{new})$\ \tcp*{Compute semantic vector}
                
                $\mathcal{B}_\text{sim} \leftarrow \text{Top-}K_\text{dedup}(\mathbf{v}_\text{new}, \mathcal{B})$\ \tcp*{Retrieve top-$K_\text{dedup}$ similar set}
                
                $IsNovel \leftarrow \Phi_\text{dedup}\left( C_\text{new}, \mathcal{B}_\text{sim} \right)$\;
                
                \If{$IsNovel$ is True}{
                    $\mathcal{B} \leftarrow \mathcal{B} \cup \{ C_\text{new} \}$ \tcp*{Add novel error pattern}
                }
            }
        }
    }
}
\Return $\mathcal{B}$\;

\end{algorithm2e*}

\begin{algorithm2e*}[t]
\caption{Online Procedure: Rubric Retrieval, Auditing, and Tri-State Gating}
\label{alg:intervention}
\SetKwInOut{Input}{Input}
\SetKwInOut{Output}{Output}
\SetKwInOut{Param}{Parameters}

\Input{Agents $A_{1:N}$ in execution order, Rubric criterion bank $\mathcal{B}$, Auditing agent $\Phi_\text{audit}$, Embedding Model $M_\text{emb}$}
\Param{Max revision rounds $T_\text{max}$, Rubric size $K_\text{ctx}$, Satisfaction threshold $\tau_\text{pass}$}
\Output{Final answer $\mathcal{Y}$}

\BlankLine
\ForEach{Agent $A_i \in A_{1:N}$}{
    $o_i^{(0)} \leftarrow \Phi_i(x_i, \mathcal{R}_i, \mathcal{K}_i)$ \tcp*{Initial Generation}
    $t \leftarrow 0$\;
    
    \While{True}{
        \tcp{Step 1: Contextual Rubric Retrieval}
         $\mathcal{S}_\text{scen}^{(t)}, \mathcal{S}_\text{act}^{(t)} \leftarrow \Phi_\text{audit}(x_i, o_i^{(t)})$\ \tcp*{Extract keywords}
         $\mathbf{q}_i^{(t)} \leftarrow M_\text{emb}(\mathcal{S}_\text{scen}^{(t)} \oplus \mathcal{S}_\text{act}^{(t)})$\ \tcp*{Compute query}
         $\mathcal{B}_\text{cand}^{(t)} \leftarrow \text{Top-}K_\text{cand}(\mathbf{q}_i^{(t)}, \mathcal{B})$\ \tcp*{Coarse: Semantic matching}
        $\mathcal{B}_\text{ctx}^{(t)} \leftarrow \Phi_\text{audit}(\mathcal{B}_\text{cand}^{(t)} \mid x_i, \mathcal{R}_i, o_i^{(t)})$\ \tcp*{Fine: Assemble rubric of $\le K_\text{ctx}$ criteria}
        
        \tcp{Step 2: Criterion-Level Judgment}
        $\mathcal{F}^{(t)} \leftarrow \emptyset$\;
        \ForEach{Criterion $C_k \in \mathcal{B}_\text{ctx}^{(t)}$}{
            $(v_k^{(t)}, r_k^{(t)}) \leftarrow \Phi_\text{audit}(o_i^{(t)} \mid x_i, \mathcal{R}_i, (a_k, s_k))$\;
            \If{$v_k^{(t)} = 1$}{
                $\mathcal{F}^{(t)} \leftarrow \mathcal{F}^{(t)} \cup \{r_k^{(t)}\}$\;
            }
        }
        \tcp{Step 3: Tri-State Gating Decision}
        $p_i^{(t)} \leftarrow 1 - |\mathcal{F}^{(t)}| / |\mathcal{B}_{\text{ctx}}^{(t)}|$ \tcp*{Rubric satisfaction rate}
        \If{$p_i^{(t)} \ge \tau_{\text{pass}}$}{
            $o_i \leftarrow o_i^{(t)}$\;
            \textbf{break} \tcp*{\textbf{Pass}: Accept and broadcast}
        }
        \ElseIf{$t < T_\text{max}$}{
            $o_i^{(t+1)} \leftarrow \Phi_i(x_i, \mathcal{R}_i, \mathcal{K}_i, \mathcal{F}^{(t)})$ \tcp*{\textbf{Revise}: Regenerate}
            $t \leftarrow t + 1$\;
        }
        \Else{
            $o_i \leftarrow \emptyset$ \tcp*{\textbf{Withhold}: Do not broadcast}
            \textbf{break}\;
        }
    }
    \tcp{Propagate Output to Successors}
    \If{$o_i \neq \emptyset$}{
        \ForEach{Agent $A_j \in \mathcal{N}(A_i)$}{
            $\mathcal{K}_j \leftarrow \mathcal{K}_j \cup \{(\mathcal{R}_i, o_i)\}$ \tcp*{Propagate to downstream agents}
        }
    }
}

\Return $o_N$\;

\end{algorithm2e*}

\clearpage

\begin{figure*}[h]
\centering
\small
\begin{tcolorbox}[title=A Rubric Criterion Example]
Name: COVERAGE\_PERCENTAGE\_MISUNDERSTANDING\\
Diagnostic Definition: This error involves misunderstanding the calculation of the final amount paid without insurance, due to confusion in handling the interaction between percentage coverage and total cost.\\
Applicability Condition: When the problem involves percentage coverage of costs and requires the computation of savings based on the percentage covered.\\
Inspection Directive: Attention! Check if the total procedure cost was considered correctly before applying the percentage to compute savings.\\
\end{tcolorbox}
\caption{An example criterion from the constructed bank for the math domain. The \textit{Diagnostic Definition} and \textit{Applicability Condition} drive retrieval and selection, while the \textit{Applicability Condition} and \textit{Inspection Directive} are the fields handed to the auditing agent at judgment time.}
\label{fig:indicator_example}
\end{figure*}

\begin{figure*}[h]
\centering
\small
\begin{tcolorbox}[title=Generic Criterion for Math]
Name: CRITICAL\_MATH\_LOGIC\_AUDIT\\
Diagnostic Definition: A focused audit to detect substantive logical fallacies, calculation errors, or conditional oversights that invalidate the final result.\\
Applicability Condition: The Agent is performing mathematical reasoning, derivation, or calculation.\\
Inspection Directive: You are an Objective Math Auditor. Your duty is to **verify** the agent's logic, not to rewrite their solution.\\
**Audit Standards:**\\
1. **Fatal Errors ONLY**: Flag specific steps that are mathematically FALSE. Do not critique efficiency, style, or 'better methods'. If the logic holds, let it pass.\\
2. **Verify, Don't Assume**: Don't just look at the answer. Check if the intermediate deductions actually support the conclusion.\\\\
**Potential Risk Areas to Scan (Heuristics):**\\
- **Hallucinations**: Using non-existent theorems or making up numbers.\\
- **Logic Gaps**: Jumping to conclusions without proof (e.g., assuming symmetry/maximums).\\
- **Boundary Neglect**: Missing edge cases (zero, negative, empty sets) or necessary/sufficient conditions.\\
- **Calculation Failures**: Basic arithmetic errors that propagate to the final result.\\
3. Be careful not to ask this agent to do things outside of their responsibilities, just to see if what they are doing is right or wrong\\
\end{tcolorbox}
\caption{The generic criterion designed for the math domain.}
\label{fig:indicator_math}
\end{figure*}

\begin{figure*}[h]
\centering
\small
\begin{tcolorbox}[title=Prompt for Math Auditor]
You are an Objective Logic Auditor.\\
Your task is to verify if a specific team member (**Agent Role**) has committed a **FATAL LOGIC ERROR** regarding a specific **Area of Concern**.\\

\#\#\# The "Impact \& Action" Protocol\\
1. **Presumption of Validity**: You must assume the Agent's reasoning is correct unless you find irrefutable evidence of a fatal flaw.\\
2. **The "Actionability" Test**: If you cannot provide a specific, mathematical correction (a formula, a step, or a value), **IT IS NOT A FLAW**.\\
3. **The "Impact" Test**: If the Agent's phrasing is imperfect but the **FINAL ANSWER** remains mathematically correct, **IT IS NOT A FLAW**.\\

\#\#\# Judgment Criteria\\
**[Area of Concern]**: Context: \{applicability\_condition\}\\
Specific Risk: \{inspection\_directive\}\\

---\\

\#\#\# CONTEXT\\
- **Task**: \{task\}\\
- **Agent Role**: \{role\}\\
- **Agent Output**: \{agent\_output\}\\

---\\

\#\#\# OUTPUT FORMAT (JSON ONLY)\\
You must generate the fields in this **EXACT ORDER**. The logical flow determines the verdict.\\

\{\\
    "evidence\_quote": "Verbatim quote of the problematic part. Write 'N/A' if valid.",\\
    "analysis": "Explain WHY this specific part violates the Area of Concern. Focus on logic, not style. Try to express in a concise and to the point manner, avoid lengthy speeches. Write 'N/A' if valid.",\\
    "suggestion": "Concrete instruction on how to fix it (e.g., 'Change x to y', 'Apply formula Z'). If no fix is needed or possible, write 'N/A'.",\\
    "impact\_assessment": "Simulate the correction. Does the FINAL ANSWER or core conclusion change? (YES/NO) and brief reason.",\\
    "is\_flawed": boolean // Set to true ONLY if 'suggestion' is concrete AND 'impact\_assessment' is YES. Otherwise false.\\
\}
\end{tcolorbox}
\caption{The prompt template for the auditing agent in the math domain. The \textit{Area of Concern} block is instantiated per criterion: the applicability condition and the inspection directive are concatenated and substituted into the slot.}
\label{fig:prompt_math}
\end{figure*}

\begin{figure*}[h]
\centering
\small
\begin{tcolorbox}[title=Prompt for Code Auditor]
You are a Senior Code Auditor and Architect.\\
Your task is to verify if a specific team member (**Agent Role**) has committed a **FATAL CODING ERROR** regarding a specific **Area of Concern**.\\

\#\#\# The "Impact \& Action" Protocol
1. **Presumption of Validity**: You must assume the Agent's code is functionally correct unless you find irrefutable evidence of a fatal flaw (syntax error, logic bug, or interface violation).\\
2. **The "Actionability" Test**: If you cannot provide a specific code correction (a line change, a logic fix, or a parameter adjustment), **IT IS NOT A FLAW**.\\
3. **The "Impact" Test**: If the code is inefficient, verbose, or stylistically non-standard but **EXECUTES CORRECTLY** and returns the right result, **IT IS NOT A FLAW**.\\

\#\#\# Judgment Criteria\\
**[Area of Concern]**: Context: \{applicability\_condition\}\\
Specific Risk: \{inspection\_directive\}\\

---\\

\#\#\# CONTEXT\\
- **Task**: \{task\}\\
- **Agent Role**: \{role\}\\
- **Agent Output**: \{agent\_output\}\\

---\\

\#\#\# OUTPUT FORMAT (JSON ONLY)
You must generate the fields in this **EXACT ORDER**. The logical flow determines the verdict.\\

\{\\
    "evidence\_quote": "Verbatim quote of the problematic code snippet. Write 'N/A' if valid.",\\
    "analysis": "Explain WHY this specific part violates the Area of Concern. Focus on functional correctness (bugs/crashes), not style (PEP8/comments).Try to express in a concise and to the point manner, avoid lengthy speeches. Write 'N/A' if valid.",\\
    "suggestion": "Concrete instruction on how to fix the code (e.g., 'Change index i to i+1', 'Import module X'). If no fix is needed, write 'N/A'.",\\
    "impact\_assessment": "Simulate the correction. Does it fix a runtime error, infinite loop, or incorrect output? (YES/NO) and brief reason.",\\
    "is\_flawed": boolean // Set to true ONLY if 'suggestion' is concrete AND 'impact\_assessment' is YES. Otherwise false.\\
\}
\end{tcolorbox}
\caption{The prompt template for the auditing agent in the code domain.}
\label{fig:prompt_code}
\end{figure*}

\begin{figure*}[h]
\centering
\small
\begin{tcolorbox}[title=Prompt for Miner, fontupper=\scriptsize]
You are an AI acting as a **Lead Mathematics Auditor and Logic Specialist**, specifically optimized for the MATH dataset (high-difficulty competitions like AMC, AIME).

\#\#\# Background \& Goal\\
**Background**: An agent team has attempted to solve a complex math problem, and **the team's final answer is INCORRECT**.\\
**Goal**: Synthesize the known problem (`problem`), standard solution (`solution`), and the Agent's `output` to strictly evaluate the Agent's reasoning process.\\

**IMPORTANT CONTEXT**: \\
The provided `solution` is a standard, single-path reference answer. However, the Agent is part of a Multi-Agent System (MAS). \\
- Its `output` depends on its `agent\_role` (e.g., a "Python Coder" writes code, a "Critic" critiques). \\
- **DO NOT** penalize the Agent simply because its output does not look like the standard `solution` (e.g., using code instead of pure derivation is valid if the role permits).\\
- Only penalize **logical errors**, **calculation errors**, or **hallucinations** that contradict mathematical truths.\\

\#\#\# MATH Input Context\\
1. **`problem`**: \\
\{problem\}\\
2. **`solution`**: (Ground Truth)\\
\{solution\}\\
3. **`agent\_role`**: \\
\{agent\_role\}\\
4. **`output`**: (Agent's Attempt)\\
\{output\}\\

\#\#\# Phase 1: Diagnosis\\
Please execute the following logical judgment:\\
1. Assess whether the Agent's output is logically and mathematically correct **within the scope of its role**.\\
2. **AUDIT STRATEGY (CRITICAL)**:\\
   - **DO NOT STOP at the first error.** You must scan the ENTIRE output line by line.\\
   - Independent errors often exist (e.g., a logical fallacy in Step 1 AND a formatting error in the Final Answer).\\
   - You are expected to find **MULTIPLE distinct errors** (less than 5) if they exist.\\
3. **Decision**:\\
   - If the output contains **NO errors**: Output `NO\_ERROR`.\\
   - If the output contains **errors**: Identify **ALL** of them and proceed to Phase 2.\\

\#\#\# Phase 2: Metric Extraction\\
Transform **EACH identified error** separately into a **generalized** JSON metric object.\\
**CRITICAL**: The `name`, `detailed\_definition`, `trigger\_condition`, and `example\_error` must be **generalizable** to other similar math problems.\\

1. **`name`**:\\
    *   **Requirement**: Summarize the error pattern. It can be **appropriately longer** to avoid ID collisions.\\
    *   **Format**: `UPPER\_CASE\_WITH\_UNDERSCORES`.\\
2. **`domain\_tag`**:\\
    *   **Requirement**: Classify this error into a specific mathematical or operational domain.\\
    *   **Examples**: "Geometry", "Probability", "Algebra", "Number Theory", "Python Implementation", "Logical Reasoning".\\
3. **`detailed\_definition`**:\\
    *   **Requirement**: Define the **ROOT CAUSE** or **Mental Misconception** behind the error. Do not just say "used wrong formula"; explain "confused concept A with concept B".\\
    *   **Format**: "This error occurs when the agent [misconception], leading to [consequence]."\\
4. **`evaluator\_prompt`**:\\
    Contains the trigger condition for retrieving this metric:\\

    *   **`trigger\_condition`**:\\
        *   **Requirement**: Describe the **Context** or **Action** where this error is likely to happen. **DO NOT** assume the error has already occurred (Decriminalized).\\
        *   **Format**: "When the problem involves [context]..." OR "When the agent attempts to [action]..."\\
5. **`example\_error`**:\\
    *   **Requirement**: Provide a concrete example of the error AND the logic for why it is wrong/how to fix it.\\
    *   **Format**: "Error Snippet: [Quote agent's wrong step] | Correction Logic: [Explain why it is wrong and what the correct approach/formula should be]."\\

\#\#\# Output Format\\
- If no error: Output `NO\_ERROR` only.\\
- If errors exist: **ALWAYS Output a JSON LIST** containing one or more metric objects.\\
  - Structure: `[ \{ "name": "ERROR\_1", ... \}, \{ "name": "ERROR\_2", ... \} ]`\\
- **CRITICAL JSON SYNTAX RULE**:\\
  - When writing LaTeX inside JSON strings, **YOU MUST DOUBLE-ESCAPE BACKSLASHES**.\\
  - **WRONG**: `"equation": "\\frac\{1\}\{2\}"` (This causes JSON parse error!)\\
  - **CORRECT**: `"equation": "\\\\frac\{1\}\{2\}"` (This works!)\\
\end{tcolorbox}
\caption{The prompt template for the rubric miner during offline bank construction.}
\label{fig:prompt_teacher}
\end{figure*}

\begin{figure*}
    \centering
\begin{tcolorbox}[notitle, colback=white, colframe=softblue,
       fontupper=\small,
       boxrule=3pt, boxsep=0.5pt, enhanced, 
       shadow={3pt}{-3pt}{0pt}{opacity=1,mygray},
       title={\large \includegraphics[height=22pt]{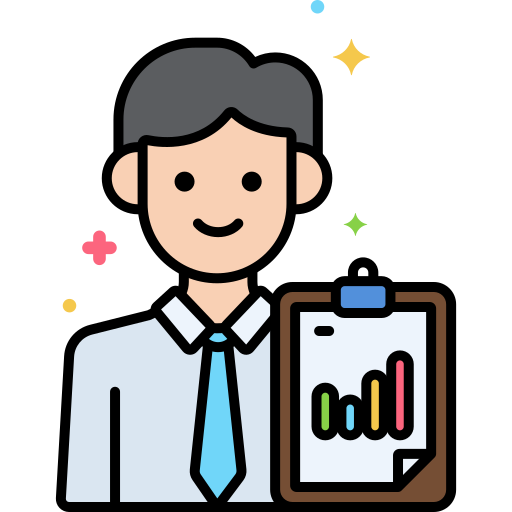} {\large Math Solver: Revision Round 2/2 Output $o^{(2)}$}}]
The solver accepts the second audit's feedback and revises the geometry. It now
states that the internal angle of the nonagon does not control the rounded
corner in the offset boundary. Instead, each corner arc is determined by the
external angle of the regular nonagon.\\

Corrected calculation:\\

1. External angle of the nonagon:

\[
\frac{2\pi}{9}.
\]

2. Arc length for each rounded corner, with radius \(1\):

\[
1\cdot\frac{2\pi}{9}=\frac{2\pi}{9}.
\]

3. Total rounded-corner contribution across all nine vertices:

\[
9\cdot\frac{2\pi}{9}=2\pi.
\]

4. Straight-side contribution:

\[
9\times 2=18.
\]

Thus the corrected perimeter is

\[
18+2\pi,
\]

and the solver outputs

\[
\boxed{18+2\pi}.
\]
\end{tcolorbox}
    \caption{The output of the math solver agent in Revision Round 2.}
    \label{fig:case_reasoner_2}
\end{figure*}

\begin{figure*}
    \centering
\begin{tcolorbox}[notitle, colback=white, colframe=softorange,
       fontupper=\small,
       boxrule=3pt, boxsep=0.5pt, enhanced, 
       shadow={3pt}{-3pt}{0pt}{opacity=1,mygray},
       title={\large \includegraphics[height=22pt]{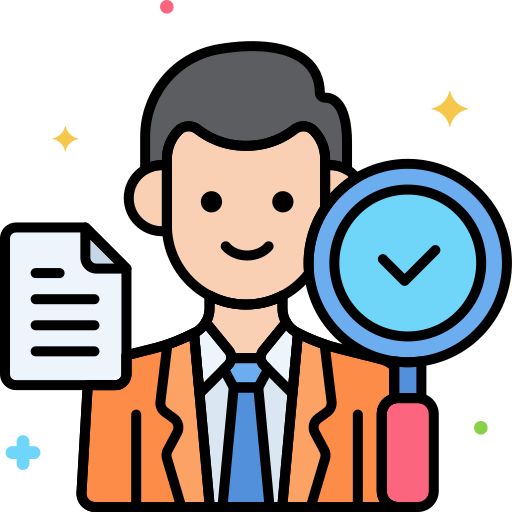} {\large Auditor: Revision Round 2/2 Evaluation}}]
[1] Matched Criteria:

\quad \quad - CALCULATION\_CONTINGENT\_ON\_ANGLES

\quad \quad - INCORRECT\_SUMMATION\_OF\_SEGMENT\_LENGTHS

\quad \quad - GEOMETRIC\_COMPOUND\_CALCULATION\_CHECK

\quad \quad - ANGLE\_DEPENDENCY\_CHECK

\quad \quad - ANGLE\_TYPE\_IDENTIFICATION\\

[2] Evaluations:

\quad \quad - CALCULATION\_CONTINGENT\_ON\_ANGLES: Correct

\quad \quad - INCORRECT\_SUMMATION\_OF\_SEGMENT\_LENGTHS: Correct

\quad \quad - GEOMETRIC\_COMPOUND\_CALCULATION\_CHECK: Correct

\quad \quad - ANGLE\_DEPENDENCY\_CHECK: Correct

\quad \quad - ANGLE\_TYPE\_IDENTIFICATION: Correct\\

[3] Pass Count:

5/5, threshold 60\%, action: passed.\\

[4] Feedback:

[EMPTY]
\end{tcolorbox}
    \caption{The auditor's judgments and feedback to the Round 2 output.}
    \label{fig:case_rectifier_2}
\end{figure*}

\end{document}